\documentclass[letterpaper, 10 pt, conference]{ieeeconf}  % Comment this line out if you need a4paper

\newcommand{\debugcommand}[1]{}        % empty for final
\newcommand{\finalcommand}[1]{#1}      % #1    for final

\IEEEoverridecommandlockouts                              % This command is only needed if 
\usepackage[utf8]{inputenc}
\usepackage{graphicx}
\usepackage{amsmath}
\usepackage{amssymb}
\usepackage{textcomp}
\usepackage{graphicx}
\usepackage[caption=false]{subfig}
\usepackage{color}
\usepackage{ifpdf}
\usepackage{url}
\usepackage{enumerate}
\usepackage{array}
\usepackage[percent]{overpic}
\usepackage{tikz}
\usetikzlibrary{bayesnet}
\usetikzlibrary{shapes.geometric,calc}
\tikzstyle{filledcircle} = [circle, draw=black, inner sep=1pt, outer sep=0pt]
\tikzstyle{filledcircle15} = [circle, draw=black, inner sep=1pt, outer sep=0pt, minimum size=15pt]

\newcommand{\Mtikzfigoutdoordataset}{
    \begin{tikzpicture}[]
        \footnotesize
        \begin{scope}[shift={(-3.0,0)}]
            \draw[dashed] (0,0) circle(3.0);
            \draw[dashed] (0,0) -- (  6:3.0) node[filledcircle15, fill=red!20] (tube) {$A_{15}$};
            \draw[dashed] (0,0) -- ( 30:3.0) node[filledcircle15, fill=red!20] (tube) {$A_{14}$};
            \draw[dashed] (0,0) -- ( 54:3.0) node[filledcircle15, fill=red!20] (tube) {$A_{13}$};
            \draw[dashed] (0,0) -- ( 78:3.0) node[filledcircle15, fill=red!20] (tube) {$A_{12}$};
            \draw[dashed] (0,0) -- (102:3.0) node[filledcircle15, fill=red!20] (tube) {$A_{11}$};
            \draw[dashed] (0,0) -- (126:3.0) node[filledcircle15, fill=red!20] (tube) {$A_{10}$};
            \draw[dashed] (0,0) -- (150:3.0) node[filledcircle15, fill=red!20] (tube) {$A_{09}$};
            \draw[dashed] (0,0) -- (174:3.0) node[filledcircle15, fill=red!20] (tube) {$A_{08}$};
            \draw[dashed] (0,0) -- (198:3.0) node[filledcircle15, fill=red!20] (tube) {$A_{07}$};
            \draw[dashed] (0,0) -- (222:3.0) node[filledcircle15, fill=red!20] (tube) {$A_{06}$};
            \draw[dashed] (0,0) -- (246:3.0) node[filledcircle15, fill=red!20] (tube) {$A_{05}$};
            \draw[dashed] (0,0) -- (270:3.0) node[filledcircle15, fill=red!20] (tube) {$A_{04}$};
            \draw[dashed] (0,0) -- (294:3.0) node[filledcircle15, fill=red!20] (tube) {$A_{03}$};
            \draw[dashed] (0,0) -- (318:3.0) node[filledcircle15, fill=red!20] (tube) {$A_{02}$};
            \draw[dashed] (0,0) -- (342:3.0) node[filledcircle15, fill=red!20] (tube) {$A_{01}$};
            \node[filledcircle, fill=red!20] (tube) {$\Mtube_{A}$};
        \end{scope}
        \begin{scope}[shift={( 3.0,0)}]
            \draw[dashed] (0,0) circle(3.0);
            \draw[dashed] (0,0) -- ( 18:3.0) node[filledcircle15, fill=blue!20] (tube) {$B_{09}$};
            \draw[dashed] (0,0) -- ( 42:3.0) node[filledcircle15, fill=blue!20] (tube) {$B_{10}$};
            \draw[dashed] (0,0) -- ( 66:3.0) node[filledcircle15, fill=blue!20] (tube) {$B_{11}$};
            \draw[dashed] (0,0) -- ( 90:3.0) node[filledcircle15, fill=blue!20] (tube) {$B_{12}$};
            \draw[dashed] (0,0) -- (114:3.0) node[filledcircle15, fill=blue!20] (tube) {$B_{13}$};
            \draw[dashed] (0,0) -- (138:3.0) node[filledcircle15, fill=blue!20] (tube) {$B_{14}$};
            \draw[dashed] (0,0) -- (162:3.0) node[filledcircle15, fill=blue!20] (tube) {$B_{15}$};
            \draw[dashed] (0,0) -- (186:3.0) node[filledcircle15, fill=blue!20] (tube) {$B_{01}$};
            \draw[dashed] (0,0) -- (210:3.0) node[filledcircle15, fill=blue!20] (tube) {$B_{02}$};
            \draw[dashed] (0,0) -- (234:3.0) node[filledcircle15, fill=blue!20] (tube) {$B_{03}$};
            \draw[dashed] (0,0) -- (258:3.0) node[filledcircle15, fill=blue!20] (tube) {$B_{04}$};
            \draw[dashed] (0,0) -- (282:3.0) node[filledcircle15, fill=blue!20] (tube) {$B_{05}$};
            \draw[dashed] (0,0) -- (306:3.0) node[filledcircle15, fill=blue!20] (tube) {$B_{06}$};
            \draw[dashed] (0,0) -- (330:3.0) node[filledcircle15, fill=blue!20] (tube) {$B_{07}$};
            \draw[dashed] (0,0) -- (354:3.0) node[filledcircle15, fill=blue!20] (tube) {$B_{08}$};
            \node[filledcircle, fill=blue!20] (tube) {$\Mtube_{B}$};
        \end{scope}
        \draw[<->] (-6.0, -3.40) -- ( 6.0, -3.40);
        \node[] at (0., -3.20) (label6m) {\normalsize \SI{6}{\meter}};
    \end{tikzpicture}
}

\usepackage{balance}
\usepackage{multirow,booktabs}
\usepackage{hyperref}
\usepackage{siunitx}
\let\labelindent\relax % \labelindent redefined in enumitem
\usepackage[inline]{enumitem}
\usepackage{csquotes}
\usepackage[export]{adjustbox}
\usepackage{nicefrac}

\usepackage{algpseudocode,algorithm,algorithmicx}
\newcommand*\Let[2]{\State #1 $\gets$ #2}
\algrenewcommand\algorithmicrequire{\textbf{Precondition:}}
\algrenewcommand\algorithmicensure{\textbf{Postcondition:}}

\definecolor{nicered}{rgb}{0.8,0.2,0.2}
\definecolor{niceblu}{rgb}{0.2,0.2,0.8}
\definecolor{nicegrn}{rgb}{0.2,0.6,0.2}
\definecolor{niceyel}{rgb}{0.8,0.8,0.1}
\definecolor{nicegry}{rgb}{0.5,0.5,0.5}
\definecolor{hrcolor}{rgb}{0.3,0.3,0.3}

\newcounter{debugcounter}
\newcommand{\debug}[1]{%
\addtocounter{debugcounter}{1}%
{\color{nicered}\footnote[{\arabic{debugcounter}}]%
{\color{nicered}#1}}}
\newcommand{\silentcomment}[1]{}

\newcommand{\pdebug}[1]{\begin{quote}{\color{nicered}\rule{\linewidth}{0.5mm}}\\\footnotesize\sl #1\\[-1ex]\color{nicered}\rule{\linewidth}{0.5mm}\end{quote}}
\newcommand{\idebug}[1]{{\color{nicered}\footnotesize\sl #1}}
\finalcommand{\renewcommand{\debug}[1]{}}
\finalcommand{\renewcommand{\pdebug}[1]{}}
\finalcommand{\renewcommand{\idebug}[1]{}}

\ifpdf

\else

\fi

\graphicspath{{vector/}}

\newcommand{\Mmesh}{\mathcal{M}}
\newcommand{\Mmeshfilter}{\widehat{\Mmesh}}
\newcommand{\Mimageset}{\mathcal{I}}
\newcommand{\Mimage}{\mathbf{I}}
\newcommand{\Mdepth}{\mathbf{D}}
\newcommand{\Mcount}{N}
\newcommand{\Mimagecountperseverance}{\Mcount^{\Mperseverance}}
\newcommand{\Mpoint}{\mathbf{p}}
\newcommand{\Mcorrespondences}{\mathcal{C}}
\newcommand{\Mcorrespondencescount}{\Mcount^{\Mcorrespondences}}
\newcommand{\Miterationcount}{\Mcount^{\text{iter}}}
\newcommand{\Mimagepatch}{\widehat{\Mimage}}
\newcommand{\Mdepthpatch}{\widehat{\Mdepth}}

\newcommand{\Mleft}{\text{L}}
\newcommand{\Mright}{\text{R}}

\newcommand{\Mimagefetch}{\Mimage^{\Mfetch}}
\newcommand{\Mimagefetchleft}[1]{\Mimagefetch_{\Mleft{}{#1}}}
\newcommand{\Mimagefetchright}[1]{\Mimagefetch_{\Mright{}{#1}}}
\newcommand{\Mimageperseverance}{\Mimage^{\Mperseverance}}
\newcommand{\Mimageperseveranceleft}[1]{\Mimageperseverance_{\Mleft{}{#1}}}
\newcommand{\Mimageperseveranceright}[1]{\Mimageperseverance_{\Mright{}{#1}}}
\newcommand{\Mimagesynthetic}{\Mimage^{\Mvirtualviewpoint{}}}
\newcommand{\Mimagesyntheticleft}[1]{\Mimagesynthetic_{{#1}\Mleft{}}}
\newcommand{\Mimagesyntheticright}[1]{\Mimagesynthetic_{{#1}\Mright{}}}

\newcommand{\Mimagepatchsynthetic}{\Mimagepatch^{\Mvirtualviewpoint{}}}
\newcommand{\Mimagepatchsyntheticleft}{\Mimagepatchsynthetic_{\Mleft{}}}
\newcommand{\Mimagepatchsyntheticright}{\Mimagepatchsynthetic_{\Mright{}}}

\newcommand{\Mimagepatchleft}[1]{\Mimagepatch^{{#1}}_{\Mleft{}}}

\newcommand{\Mdepthpatchleft}[1]{\Mdepthpatch^{{#1}}_{\Mleft{}}}
\newcommand{\Mdepthpatchright}[1]{\Mdepthpatch^{{#1}}_{\Mright{}}}

\newcommand{\Mtransform}[2]{{}^{#1}{\mathbf{T}}_{#2}}
\newcommand{\Mtransformhat}[2]{{}^{#1}{\mathbf{T}}_{\widehat{{#2}}}}
\newcommand{\Mpointinframe}[1]{{}^{#1}{\Mpoint}}
\newcommand{\Mpointtildeinframe}[1]{{}^{#1}{\widetilde{\Mpoint}}}
\newcommand{\Mworld}{\mathcal{W}}

\newcommand{\Mvirtualviewpoint}{\mathcal{V}}
\newcommand{\Mtube}{\mathcal{T}}
\newcommand{\Mperseverance}{\text{M2020}}
\newcommand{\Mfetch}{\text{SFR}}

\newcommand{\Mcorrespondencesworldfetch}{\Mcorrespondences_{\Mworld/\Mfetch}}
\newcommand{\Mtranslationuncertainty}{\widetilde{t}}
\newcommand{\Mrotationuncertainty}{\widetilde{r}}
\newcommand{\Mdecay}{\gamma}
\newcommand{\Mpatchlength}{\ell}
\newcommand{\Mpixelcoordinateu}[1]{u_{{#1}}}
\newcommand{\Mpixelcoordinatev}[1]{v_{{#1}}}

\newcommand{\Mpixelcoordinateusynthetic}[1]{\Mpixelcoordinateu{{#1}}^{\Mvirtualviewpoint}}
\newcommand{\Mpixelcoordinatevsynthetic}[1]{\Mpixelcoordinatev{{#1}}^{\Mvirtualviewpoint}}
\newcommand{\Mpixelcoordinatessynthetic}[1]{(\Mpixelcoordinateusynthetic{{#1}}, \Mpixelcoordinatevsynthetic{{#1}})}
\newcommand{\Mpixelcoordinateufetch}[1]{\Mpixelcoordinateu{{#1}}^{\Mfetch}}
\newcommand{\Mpixelcoordinatevfetch}[1]{\Mpixelcoordinatev{{#1}}^{\Mfetch}}
\newcommand{\Mpixelcoordinatesfetch}[1]{(\Mpixelcoordinateufetch{{#1}}, \Mpixelcoordinatevfetch{{#1}})}

\newcommand{\Mmatchthreshold}{\epsilon_{u}}

\usepackage{eso-pic}

\graphicspath{{figures/}}
\DeclareGraphicsExtensions{.pdf,.jpg,.jpeg,.png}

\makeatletter
\patchcmd{\@makecaption}
  {\scshape}
  {}
  {}
  {}
\makeatletter
\patchcmd{\@makecaption}
  {\\}
  {:\ }
  {}
  {}

\title{\LARGE \bf
    Rover Relocalization for Mars Sample Return by\\%
    Virtual Template Synthesis and Matching
}

\author{Tu-Hoa Pham, William Seto, Shreyansh Daftry, Barry Ridge, Johanna Hansen,\\%
    Tristan Thrush, Mark Van der Merwe, Gerard Maggiolino, Alexander Brinkman,\\%
    John Mayo, Yang Cheng, Curtis Padgett, Eric Kulczycki and Renaud Detry\\% <-this % stops a space
Jet Propulsion Laboratory, California Institute of Technology, Pasadena, CA%
\thanks{J. Hanssen, T. Thrush, M. Van der Merwe, G. Maggiolino conducted their work as JPL interns. They are now with, respectively, McGill University, Facebook AI Research, University of Michigan, Carnegie Mellon University.}% <-this % stops a space
\thanks{The research described in this paper was carried out at the Jet Propulsion Laboratory, California Institute of Technology, under a contract with the National Aeronautics and Space Administration. The information presented about potential Mars Sample Return architectures is provided for planning and discussion purposes only. NASA has made no official decision to implement Mars Sample Return.}
\thanks{Copyright 2020 California Institute of Technology. U.S. Government sponsorship acknowledged.}% <-this % stops a space
}

\begin{document}

\maketitle
\thispagestyle{empty}
\pagestyle{empty}

%%%%%%%%%%%%%%%%%%%%%%%%%%%%%%%%%%%%%%%%%%%%%%%%%%%%%%%%%%%%%%%%%%%%%%%%%%%%%%%%
\begin{abstract}
    We consider the problem of rover relocalization in the context of the notional Mars Sample Return campaign. In this campaign, a rover (R1) needs to be capable of autonomously navigating and localizing itself within an area of approximately $\SI{50}{} \times \SI{50}{\meter}$ using reference images collected years earlier by another rover (R0). We propose a visual localizer that exhibits robustness to the relatively barren terrain that we expect to find in relevant areas, and to large lighting and viewpoint differences between R0 and R1. The localizer synthesizes partial renderings of a mesh built from reference R0 images and matches those to R1 images. We evaluate our method on a dataset totaling 2160 images covering the range of expected environmental conditions (terrain, lighting, approach angle). Experimental results show the effectiveness of our approach. This work informs the Mars Sample Return campaign on the choice of a site where Perseverance (R0) will place a set of sample tubes for future retrieval by another rover (R1).
\end{abstract}
%%%%%%%%%%%%%%%%%%%%%%%%%%%%%%%%%%%%%%%%%%%%%%%%%%%%%%%%%%%%%%%%%%%%%%%%%%%%%%%%

\section{Introduction}
\label{sec:introduction}

The Mars 2020 \emph{Perseverance} rover that launched in July will search for signs of ancient life on Mars by collecting samples from Martian rocks and soil using an arm-mounted drill. These sample will be stored in hermetically-sealed sample tubes and released at one or multiple sample cache \emph{depots}, for possible recovery via a notional NASA-ESA follow-up mission that would land in 2028,  Mars Sample Return (MSR)~\cite{aa:muirhead:2020}. The mission would include a rover (the sample-fetching rover, SFR) and a rocket (the Mars ascent vehicle, MAV). SFR would drive to the sample cache depots, pick up the tubes and bring them back to the lander for transfer and launch to Mars orbit through the MAV.
Finally, a probe would capture the container in orbit and bring it back to Earth for sample containment and analysis.
This paper focuses on the sample retrieval phase of the campaign, in particular the problem of in-depot navigation for tube pickup.

While Mars rovers intended to survive the Martian winter have used radioisotope heating units or thermoelectric generators, the base SFR design calls for neither to limit costs, and should complete its mission in a single season before shutting down forever.
Accounting for a notional \SI{10}{\kilo\meter} drive to the sample depot(s) and back to the MAV leaves only 30 sols for SFR to pick up 36 tubes.
Due to limited Earth-Mars communication windows, \emph{ground-in-the-loop} tube pickup takes a minimum 3-sol-per-tube pickup time.
SFR thus needs to retrieve tube autonomously, with minimal guidance from operators on Earth.

\begin{figure}[h]
    \centering
    \subfloat[2014-05-02 at 16:19:00.]{
        \includegraphics[width=0.23\columnwidth]{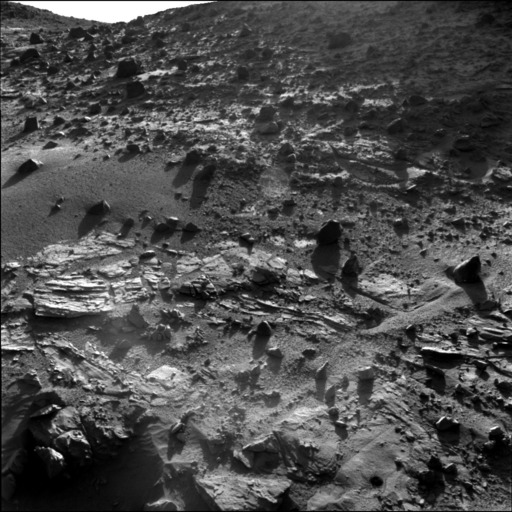}
        \includegraphics[width=0.23\columnwidth]{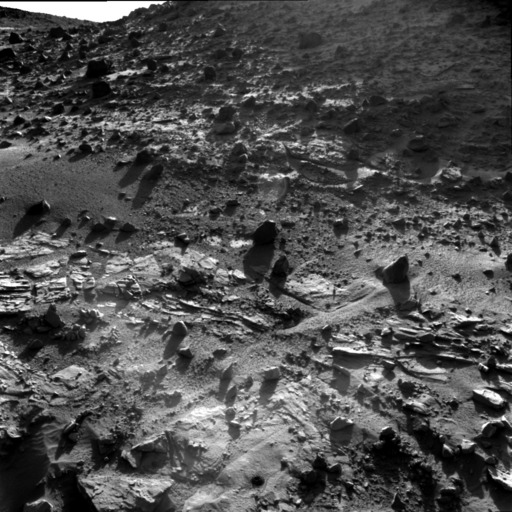}
        \label{fig:msl:p1}
    }
    \subfloat[2014-05-04 at 08:33:34.]{
        \includegraphics[width=0.23\columnwidth]{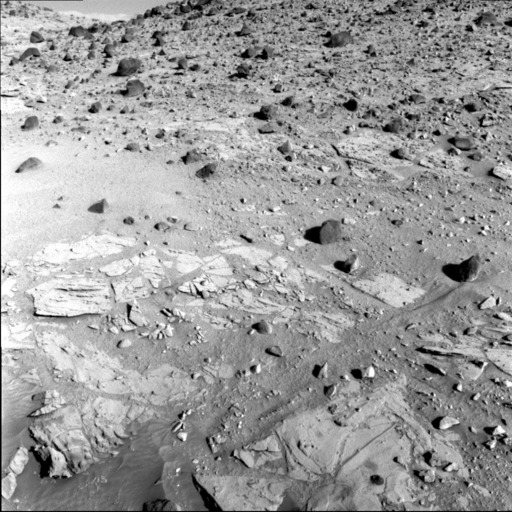}
        \includegraphics[width=0.23\columnwidth]{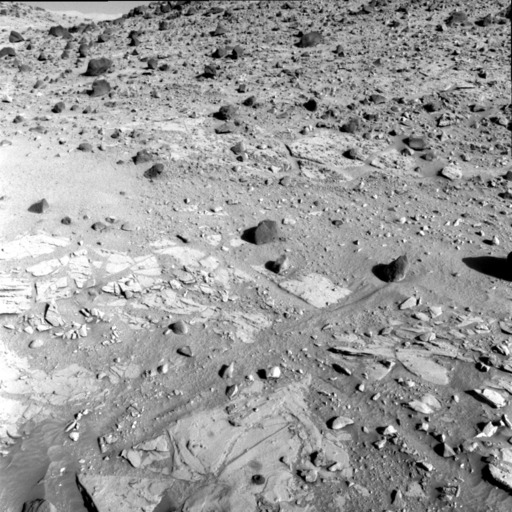}
        \label{fig:msl:p2}
    }
    \caption{Stereo images captured on Mars from the same viewpoint at different times (available \url{https://pds-imaging.jpl.nasa.gov/}).}
    \label{fig:msl}
\end{figure}

Towards this, Perseverance will document depots by capturing images as it drops sample tubes.
These images will be telemetered to Earth to reconstruct a \emph{map} of the depot annotated with tube poses.
SFR will retrieve sample tubes by \emph{relocalizing} itself with respect to that map, years later.
This is a difficult problem for multiple reasons.
In contrast to relocalization on Earth, which often benefits from human-made objects and environments with distinctive visual structures, the surface of Mars mostly consists of desert-like environments with fewer salient features.
The (changing) interaction of light with small rocks and other terrain features further complicates the problem.
The difficulty of relocalization in changing environments is well-established on Earth~\cite{cvpr:sattler:2018}.
Anecdotally, we observe the same challenges on Mars.
We depict in Fig.~\ref{fig:msl} two stereo pairs that feature-based localization~\cite{ijcv:lowe:2004} failed to align despite being from the exact same viewpoint but under different lighting.
In addition to lighting changes, relocalization must be robust to other environment changes such as dust deposition or accumulation of sand to form small drifts.
In that sense, the nature and timeline of the mission permits dedicating special care to the crafting of the map, e.g., by selecting landmarks that are more likely persist over time.

In this paper, we propose a novel method for relocalization over changing environments by Virtual Template Synthesis and Matching (VTSM).
Our work builds upon the state of the art in multiple areas of visual localization (see Section~\ref{sec:related_word}) and offers the following contributions.
\begin{itemize}
    \item A relocalization algorithm that synthesizes partial renderings of multiple points of interest on the map, as perceived from multiple virtual poses, and matches them to real observations across multiple modalities (sizes and filters) for changing environments (see Section~\ref{sec:method}).
    \item A new dataset spanning the range of environmental conditions we expect to face on Mars, including 3 terrain types, captured by 4 cameras along 60 viewpoints at 3 times of the day, totalling 2160 images annotated with ground-truth poses from a motion capture system (see Section~\ref{sec:dataset}). We release our dataset publicly to foster the research in this exciting problem for space exploration\footnote{\url{https://data.caltech.edu/records/1898}}.
    \item An extensive performance and sensitivity analysis for our method, leading to recommendations on depot mapping and navigation strategies (see Section~\ref{sec:experiments}).
\end{itemize}
We finally discuss challenges we encountered, current limitations and future extensions of our work (see Section~\ref{sec:conclusion}).

\section{Related Work}
\label{sec:related_word}

In this work, we consider the problem of \emph{relocalization} of a rover with respect to a \emph{map} built from images captured by another rover, years prior.
Autonomous localization capabilities on Mars rovers are currently limited to motion estimation by visual-inertial odometry (VIO)~\cite{jfr:maimone:2007,jfr:li:2007}, and absolute orientation estimation using gravity and Sun position in the sky~\cite{smc:ali:2005,aeroconf:lambert:2011}.
Mars rovers do not implement onboard absolute position estimation.
Instead, it is conducted on Earth, by aligning rover and orbital imagery~\cite{ar:cozman:2000,jfr:carle:2010,icarus:tao:2016} with meter-scale accuracy (Mars Reconnaissance Orbiter).
While VIO could be used to estimate relative poses between rovers, it is also vulnerable to environment changes, e.g., lighting~\cite{aeroconf:pham:2020}.

Localization techniques leveraging 3D structure~\cite{crv:bakambu:2006} can mitigate the effects of lighting changes, e.g., by performing 3D registration using iterative closest point (ICP)~\cite{pami:besl:1992} or fast point feature histograms (FPFH)~\cite{icra:rusu:2011}.
However, those tend to converge towards local optima in the absence of an accurate pose prior, especially on the rather flat terrains that may serve for depot construction.
We consider instead 3D (depth) reconstructed together with texture (color), e.g., through Structure from Motion (SfM)~\cite{cvpr:schonberger:2016} or Simultaneous Localization and Mapping (SLAM)~\cite{tro:cadena:2016}, which can accomodate maps built over multiple sessions under the same lighting~\cite{ral:giubilato:2020} or different lighting over small pose changes~\cite{icra:park:2017}.
Such 3D maps can be used for relocalization by correspondence search between features from 2D images to relocalize and a database of features associated to 3D map points~\cite{pami:sattler:2016}.
The underlying matching schemes can also be accelerated to take advantage of multi-camera systems~\cite{icra:geppert:2019} but remain subject to similar limitations as 2D local feature matching under scene changes.
Notably, \cite{rss:mcmanus:2014} showed that image patches could be more robust to changing conditions than point features, though less accurate.
While robustness can also be attained by repeated traverses under different conditions~\cite{ijrr:churchill:2013,iros:paton:2016}, depot construction on Mars would be single-shot.
In a recent benchmark of visual localization in changing conditions~\cite{cvpr:sattler:2018}, image retrieval techniques were shown to sometimes succeed at providing a coarse pose estimate when local feature matching would fail, e.g., by augmenting a mapping database with synthetic renderings~\cite{cvpr:torii:2015}.
3D map rendering was also used for relocalization by minimizing distance metrics between real and synthetic observations, such as normalized information distance~\cite{icra:pascoe:2015}, normalized cross-correlation~\cite{aeroconf:lorenz:2017} and photometric error~\cite{icra:ok:2016}, showing some robustness moderate lighting changes.
Changing shadows were further downweighted in the image alignment pipeline of~\cite{crv:gridseth:2019} however large lighting changes remain a challenge.
We build upon these works and extend relocalization capabilities by viewpoint synthesis to extensive changes in lighting and possibly scene geometry over the years.

Instead of relocalizing images across different lighting, shadow-invariant image transformations were developed in~\cite{iros:corke:2013,icra:mcmanus:2014} assuming infinitely-narrow camera sensor responses and illumination by a single Planckian source. While the former assumption can be relaxed~\cite{icra:paton:2015}, the latter may not always hold on Mars depending on the atmosphere radiance at capture time.
Similar problems were also addressed recently using deep neural networks, e.g., to learn image relighting~\cite{icra:anoosheh:2019} and image representations that are robust to lighting changes~\cite{ral:clement:2020,ral:vonstumberg:2020,icra:piasco:2019}.
While we believe such methods will result in future breakthroughs for robust relocalization on Earth, their applicability to space exploration remains restricted by the scarcity of data for training, low-compute for space-rated hardware, limited interpretability and concerns about generalization to events unseen during training.
Still, we build our approach in such a way that it could accomodate further advances in either field.

\section{Method}
\label{sec:method}

Our goal is to enable in-depot navigation for sample tube retrieval, by estimating the 6D pose of SFR with respect to a depot map built from images taken by Perseverance several years before.
To do so, we propose a relocalization method that synthesizes partial renderings of the depot map from virtual viewpoints in the vicinity of its current pose estimate and compares those to actual observations.
In the following, matrix variables are denoted in bold and scalar in italic.

\definecolor{turquoise}{RGB}{0, 255, 255}
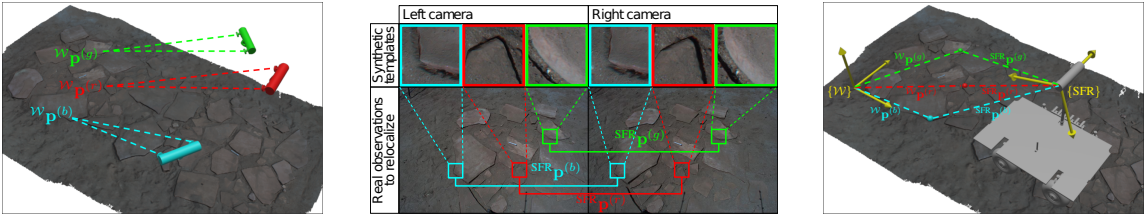
\begin{figure*}[h]
    \centering
    \resizebox{!}{0.32\columnwidth}{
        \begin{overpic}[height=0.50\columnwidth]{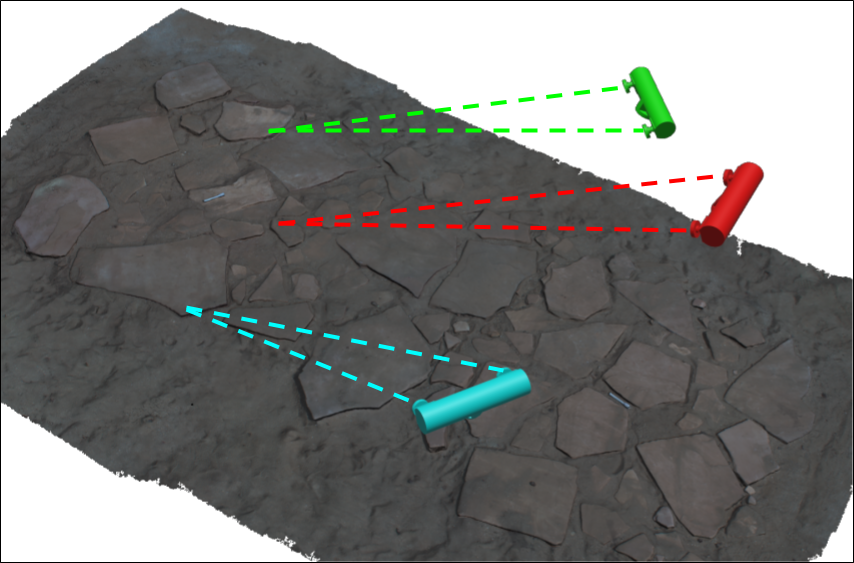}
            \put (16,47) {\color{green} $\Mpointinframe{\Mworld}^{(g)}$}
            \put (17.5,37) {\color{red} $\Mpointinframe{\Mworld}^{(r)}$}
            \put (8.5,29) {\color{turquoise} $\Mpointinframe{\Mworld}^{(b)}$}
        \end{overpic}
    }
    \quad
    \resizebox{!}{0.32\columnwidth}{
        \begin{overpic}[height=0.50\columnwidth]{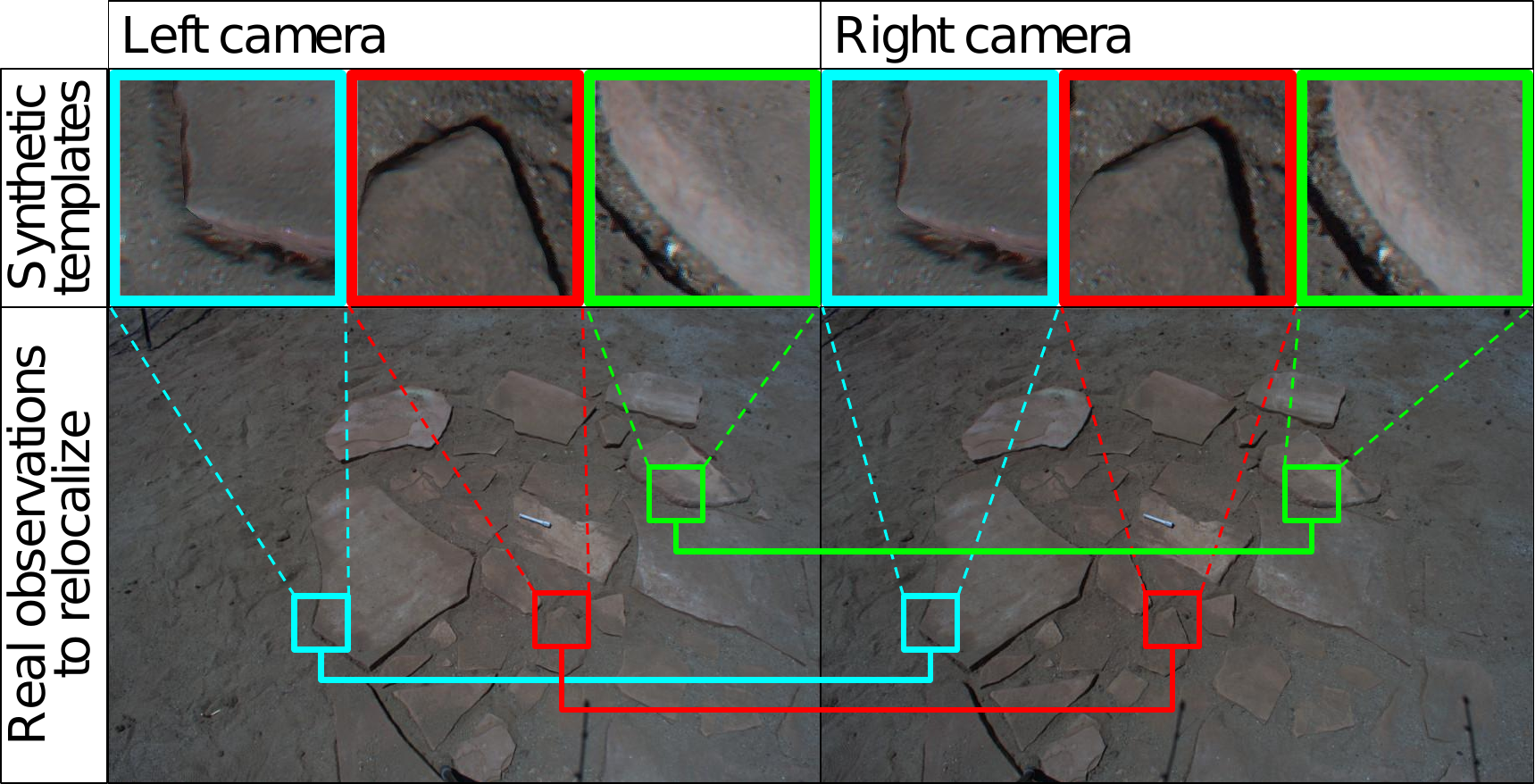}
            \put (60,17) {\color{green} $\Mpointinframe{\Mfetch}^{(g)}$}
            \put (50.5,1) {\color{red} $\Mpointinframe{\Mfetch}^{(r)}$}
            \put (39.5,9) {\color{turquoise} $\Mpointinframe{\Mfetch}^{(b)}$}
        \end{overpic}
    }
    \quad
    \resizebox{!}{0.32\columnwidth}{
        \begin{overpic}[height=0.70\columnwidth]{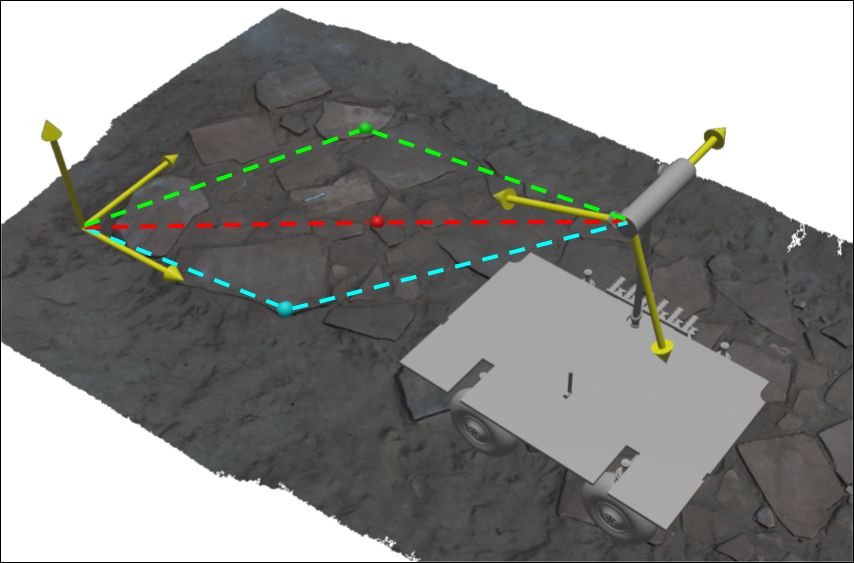}
            \put (1,37) {\color{yellow} $\{\Mworld\}$}
            \put (22,47) {\color{green} $\Mpointinframe{\Mworld}^{(g)}$}
            \put (25.5,36) {\color{red} $\Mpointinframe{\Mworld}^{(r)}$}
            \put (14.5,30) {\color{turquoise} $\Mpointinframe{\Mworld}^{(b)}$}
            \put (76,37) {\color{yellow} $\{\Mfetch\}$}
            \put (52,47) {\color{green} $\Mpointinframe{\Mfetch}^{(g)}$}
            \put (49.5,36) {\color{red} $\Mpointinframe{\Mfetch}^{(r)}$}
            \put (47.5,30) {\color{turquoise} $\Mpointinframe{\Mfetch}^{(b)}$}
        \end{overpic}
    }
    \caption{We render points $\Mpointinframe{\Mworld}$ on the mesh built from Perseverance images from multiple virtual viewpoints (left). Synthetic templates are matched to real SFR images to compute points in camera frame $\Mpointinframe{\Mfetch}$ (middle). We align both point sets to estimate the pose $\Mtransformhat{\Mworld}{\Mfetch}$ (right, conceptual rover depicted).}
    \label{fig:method}
\end{figure*}

\subsection{Overview}
\label{sec:method:overview}

Let $\Mimageset^{\Mperseverance} = (\Mimageperseveranceleft{,i}, \Mimageperseveranceright{,i})_{i\in[1,\Mimagecountperseverance]}$ denote a set of $\Mimagecountperseverance$ left and right stereo image pairs captured by Perseverance during depot construction on Mars and telemetered back to Earth.
On Earth, we register all images $\Mimageset^{\Mperseverance}$ and build a map $\Mmesh$ equiped with a global frame $\Mworld$, in which we express rover poses $\Mtransform{\Mworld}{\Mperseverance}$ as well as tubes $\Mtransform{\Mworld}{\Mtube}$.
In addition, we denote by $\Mmeshfilter$ a subset of $\Mmesh$ in which areas likely to be affected by wind (e.g., sand) have been removed on Earth between the Mars 2020 and MSR missions.

Consider now the task of relocalizing SFR several years after Perseverance.
While driving from landing site to depot, SFR's position is monitored onboard through VIO, which induces drift growing with traversed distance and corrected by manually aligning surface and orbital imagery.
When reaching the depot, relocalization with respect to $\Mmesh$ is performed by manually aligning SFR and Perseverance images instead.
Due to mission time constraints, this is done only once.
For in-depot navigation, we then assume an SFR pose estimate $\Mtransformhat{\Mworld}{\Mfetch}$ subject to VIO errors accumulated over drives post ground-in-the-loop initialization (e.g., \SI{20}{\centi\meter} for a few-meter drive, depending on specifications) and \SI{1.5}{\degree} uncertainty from onboard attitude estimation~\cite{smc:ali:2005}.

Pose alignment consists in computing a posterior pose given new camera images and a prior pose estimate $\Mtransformhat{\Mworld}{\Mfetch}$.
We compute this alignment by simulating the SFR cameras from several viewpoint hypotheses $\Mvirtualviewpoint$ and iteratively match synthetic renderings $(\Mimagesyntheticleft{}, \Mimagesyntheticright{})$ to real images $(\Mimagefetchleft{}, \Mimagefetchright{})$.
More precisely, in order to account for lighting and scene geometry changes between mapping and relocalization, we do not directly align full renderings but rather patches thereof, or \emph{synthetic templates} $(\Mimagepatchsyntheticleft{}, \Mimagepatchsyntheticright{})$, centered on a point $\Mpointinframe{\Mworld}$ and synthesized on the fly.
This enables:
\begin{enumerate*}
    \item efficiently evaluating multiple viewpoint hypotheses without having to render full images;
    \item prioritizing landmarks that are likely to remain visually similar over time, and conversely ignoring parts of the image that are likely to change;
    \item facilitating pose search as a full image may contain cast shadows (edges that do not persist with lighting changes) but small patches may be uniformly lit.
\end{enumerate*}
We match the synthetic templates to the real observations, resulting in $\Mpointinframe{\Mfetch}$, the point corresponding to $\Mpointinframe{\Mworld}$ in the SFR optical frame.
We thus collect $\Mcorrespondencescount$ correspondences $(\Mpointinframe{\Mworld}, \Mpointinframe{\Mfetch})$ to update $\Mtransformhat{\Mworld}{\Mfetch}$ by least-squares transformation estimation, and further refine it over $\Miterationcount$ iterations.
We summarize our method in Alg.~\ref{alg:overview} and Fig.~\ref{fig:method} and discuss its components in the following.

\begin{algorithm}
    \caption{Virtual Template Synthesis and Matching}
    \label{alg:overview}
    \begin{algorithmic}[1]
        \Require{$(\Mimagefetchleft{}, \Mimagefetchright{})$ stereo pair captured by SFR, $\Mmesh$ depot map, $\Mmeshfilter$ sampling mask, $\Mtransformhat{\Mworld}{\Mfetch}$ initial guess}
        \Function{Localize}{$(\Mimagefetchleft{}, \Mimagefetchright{}), \Mmesh, \Mtransformhat{\Mworld}{\Mfetch}$}
            \For{$i \gets 1 \textrm{ to } \Miterationcount$}
                \Let{$\Mcorrespondencesworldfetch$}{\{\}} \Comment{world/SFR correspondences}
                \While{$\textsc{Size}(\Mcorrespondencesworldfetch) \neq \Mcorrespondencescount$}
                \Let{$\Mtransform{\Mworld}{\Mvirtualviewpoint}$}{$\textsc{RandomizeViewpoint}(\Mtransformhat{\Mworld}{\Mfetch})$}
                    \Let{$\Mpointinframe{\Mworld}$}{$\textsc{SamplePoint}(\Mmeshfilter, \Mtransform{\Mworld}{\Mvirtualviewpoint})$}
                    \Let{$(\Mimagepatchsyntheticleft, \Mimagepatchsyntheticright)$}{$\textsc{Synthesize}(\Mmesh, \Mtransform{\Mworld}{\Mvirtualviewpoint}, \Mpointinframe{\Mworld})$}
                    \Let{$\Mpixelcoordinatesfetch{\Mleft}$}{$\textsc{Match}(\Mimagepatchsyntheticleft, \Mimagefetchleft{})$} \Comment{Left}
                    \Let{$\Mpixelcoordinatesfetch{\Mright}$}{$\textsc{Match}(\Mimagepatchsyntheticright, \Mimagefetchright{})$} \Comment{Right}
                    \If{$\textsc{Valid}(\Mpixelcoordinateufetch{\Mleft}, \Mpixelcoordinatevfetch{\Mleft}, \Mpixelcoordinateufetch{\Mright}, \Mpixelcoordinatevfetch{\Mright})$}
                        \Let{$\Mpointinframe{\Mfetch}$}{$\textsc{Stereo}(\Mpixelcoordinateufetch{\Mleft}, \Mpixelcoordinatevfetch{\Mleft}, \Mpixelcoordinateufetch{\Mright}, \Mpixelcoordinatevfetch{\Mright})$}
                        \State $\textsc{Append}(\Mcorrespondencesworldfetch, (\Mpointinframe{\Mworld}, \Mpointinframe{\Mfetch}))$
                    \EndIf
                \EndWhile
                \Let{$\Mtransformhat{\Mworld}{\Mfetch}$}{$\textsc{GetTransform}(\Mcorrespondencesworldfetch)$}
            \EndFor
            \State
            \Return{$\Mtransformhat{\Mworld}{\Mfetch}$}
        \EndFunction
    \end{algorithmic}
\end{algorithm}

\subsection{Viewpoint Sampling}
\label{sec:method:viewpoint}

Pose search begins with an initial viewpoint estimate $\Mtransformhat{\Mworld}{\Mfetch}$, which is typically computed from the previous pose and the rover's VIO-derived motion since the last alignment. We account for uncertainty incurred by VIO by searching for a best-matching pose in the neighborhood of $\Mtransformhat{\Mworld}{\Mfetch}$. Rover localization is largely a planar problem, and for perfectly flat terrain our search should be constrained to a planar search. However, as we expect substantial relief through a depot, we conduct a full 6D search.
We search the neighborhood of the initial pose by applying a perturbation transformation $\Mtransform{\Mfetch}{\Mvirtualviewpoint}$ characterized by translation and rotation search bounds $\Mtranslationuncertainty$, $\Mrotationuncertainty$ (initially, $\Mtranslationuncertainty = \Mtranslationuncertainty_0 = \SI{20}{\centi\meter}$, $\Mrotationuncertainty = \Mrotationuncertainty_0 = \SI{1.5}{\degree}$).
We build the rotational perturbation by randomly sampling a rotation axis from the unit sphere and a rotation angle from a uniform distribution on $[-\Mrotationuncertainty, \Mrotationuncertainty]$.
Similarly, the perturbation's translational  magnitude is sampled within $[-\Mtranslationuncertainty, \Mtranslationuncertainty]$ and its direction within the local surface tangent for the first iteration (planar search), then the 3D unit sphere for subsequent refinement.

\subsection{Virtual Template Synthesis}
\label{sec:method:virtualtemplatesynthesis}

In the following, we consider the depot map $\Mmesh$ as a textured polygon mesh.
Using OpenGL, we build a rendering environment reproducing SFR rectified stereo calibration parameters, enabling the synthesis of images similar in appearance to what would be captured by the real rover at arbitrary viewpoints $\Mtransform{\Mworld}{\Mvirtualviewpoint}$.
To evaluate multiple viewpoint hypotheses efficiently, we do not render full stereo images (e.g., $5472 \times 3648 \approx \SI{20}{Mpixel}$) but instead square patches of side length $\Mpatchlength$ centered on points of interest (e.g., $\Mpatchlength = 256$, about $300 \times$ smaller).
The synthesis pipeline is as follows.
Given a sampling mask $\Mmeshfilter$ of the depot areas that can reliably be used for relocalization (e.g., by keeping large rocks and filtering out sand), we randomly select a 3D point $\Mpointtildeinframe{\Mworld}$ from the vertices constituting $\Mmeshfilter$ visible from $\Mtransform{\Mworld}{\Mvirtualviewpoint}$.
When the scene geometry is not expected to change, we set $\Mmeshfilter := \Mmesh$.
We then project $\Mpointtildeinframe{\Mworld}$ into the left virtual camera frame, yielding 2D pixel coordinates $\Mpixelcoordinatessynthetic{\Mleft}$.
We thus synthesize a left template $\Mimagepatchleft{\Mmesh}$ centered on $\Mpixelcoordinatessynthetic{\Mleft}$, together with an associated depth map $\Mdepthpatchleft{\Mmesh}$.
We use the full mesh $\Mmesh$ for rendering, which contains areas that may change over time but can still be used for matching.
We estimate which pixels to keep from the synthetic template $\Mimagepatchleft{\Mmesh}$ in two ways.
First, we render the depth map $\Mdepthpatchright{\Mmeshfilter}$ associated to the sampling mask $\Mmeshfilter$ and keep all pixels of $\Mimagepatchleft{\Mmesh}$ that have a depth value in $\Mdepthpatchright{\Mmeshfilter}$ as they correspond to vertices of $\Mmeshfilter$.
Second, we identify pixels that may change locally (e.g., sand) but still serve as contrasting background for features in the foreground (e.g., rock edges).
We define those as pixels of $\Mimagepatchleft{\Mmesh}$ that do \emph{not} have a depth value in $\Mdepthpatchleft{\Mmeshfilter}$ but are beyond edges marking depth discontinuity in $\Mdepthpatchleft{\Mmesh}$.

We synthesize the right template $\Mimagepatchsyntheticright$ by similarly generating color and depth patches from a virtual \emph{right} camera.

\subsection{Synthetic-to-Real Template Matching}
\label{sec:method:matching}

We now search for $\Mpointinframe{\Mfetch}$, the point corresponding to $\Mpointinframe{\Mworld}$ in the SFR camera frame, by separately searching for left and right templates $\Mimagepatchsyntheticleft{}, \Mimagepatchsyntheticleft{}$ in the real SFR images $\Mimagefetchleft{}, \Mimagefetchright{}$, yielding SFR pixel coordinates $\Mpixelcoordinatesfetch{\Mleft}, \Mpixelcoordinatesfetch{\Mright}$ of maximum normalized cross-correlation.
We make use of the epipolar constraint between rectified stereo images to reject matches such that the vertical difference $|\Mpixelcoordinateufetch{\Mleft} - \Mpixelcoordinateufetch{\Mright}|$ exceeds a chosen threshold $\Mmatchthreshold$.
If $|\Mpixelcoordinateufetch{\Mleft} - \Mpixelcoordinateufetch{\Mright}| \leq \Mmatchthreshold$, we calculate $\Mpointinframe{\Mfetch}$ using the average vertical coordinate $0.5 * (\Mpixelcoordinateufetch{\Mleft} + \Mpixelcoordinateufetch{\Mright})$ and horizontal disparity $(\Mpixelcoordinatevfetch{\Mleft} - \Mpixelcoordinatevfetch{\Mright})$.

We exploit two strategies to facilitate synthetic-to-real matching.
First, we search for matches using different variations of synthetic templates $(\Mimagepatchsyntheticleft{}, \Mimagepatchsyntheticright{})$: size (e.g., $\nicefrac{\Mpatchlength}{2}$-length sub-template) and derivative order (e.g., direct grayscale, or processed through Sobel, Laplacian operators).
Smaller templates are easier to match across larger viewpoint differences at the cost of more false positives to filter out, while differentiating templates partially mitigates the effects of lighting differences.
Second, rather than searching for templates $(\Mimagepatchsyntheticleft{}, \Mimagepatchsyntheticright{})$ in the full SFR images $(\Mimagefetchleft{}, \Mimagefetchright{})$, we compute bounds on their possible pixel coordinates based on the pose uncertainty $\Mtranslationuncertainty$, $\Mrotationuncertainty$ and perform the template search on these sub-images (e.g., $800 \times 800$ patch within $5472 \times 3648$).

\subsection{Pose Update}
\label{sec:method:pose}

We repeat the steps described in Sections~\ref{sec:method:viewpoint} to~\ref{sec:method:matching} until reaching a target number $\Mcorrespondencescount$ of world-camera point correspondence candidates $(\Mpointinframe{\Mworld}, \Mpointinframe{\Mfetch})$ and estimate a transformation using the Umeyama algorithm~\cite{pami:umeyama:1991} and RANSAC for outlier rejection~\cite{ca:fischler:1981} within the Point Cloud Library~\cite{icra:rusu:2011}.
If successful, we set the resulting maximum-inlier transformation as new pose estimate $\Mtransformhat{\Mworld}{\Mfetch}$ and repeat the process for $\Miterationcount$ iterations or until the pose update converges within a chosen threshold (e.g., \SI{1}{\milli\meter}).
We propose additional mechanisms to facilitate the pose search.

When a transformation cannot be estimated from the correspondence candidates collected at this iteration:
\begin{itemize}
    \item \textsc{Stall}: get new correspondences from the same pose
    \item \textsc{Reseed}: apply a random perturbation to the current pose before collecting new correspondences
\end{itemize}
The \textsc{Stall} procedure is well suited when we already have at least one successful iteration, i.e., the current pose estimate is already close to the real pose.
In contrast, \textsc{Reseed} lets us evaluate multiple guesses within a potentially large initial uncertainty range (e.g., \SI{50}{\centi\meter}) while maintaining smaller synthetic viewpoint variations (e.g., \SI{20}{\centi\meter}) for local search.

When a transformation is successfully estimated:
\begin{itemize}
    \item \textsc{Anneal}: decrease synthetic viewpoint randomization following $\Mtranslationuncertainty := \Mdecay\Mtranslationuncertainty, \Mrotationuncertainty := \Mdecay\Mrotationuncertainty$, with $\Mdecay \in [0, 1]$
    \item \textsc{Distribute}: randomize synthetic viewpoints around multiple poses rather than the current estimate only
    \item \textsc{Reuse}: carry over a set number of correspondences (inliers) throughout successful iterations
\end{itemize}
The $\Mdecay$ parameter facilitates convergence by reducing synthetic viewpoint randomization over time, e.g., stop randomizing ($\Mdecay = 0$), halve every iteration ($\Mdecay = 0.5$), keep constant ($\Mdecay = 1$).
In \textsc{Distribute}, rather than focusing viewpoint synthesis around the maximum-inlier transformation from RANSAC, we spread it around the best pose candidates with frequency weighed by their inlier count (e.g., 3 candidates with 10, 15, 25 inliers, would have a \SI{20}{}, \SI{30}{}, \SI{50}{\percent} pick rate, respectively) to avoid local minima.
In \textsc{Reuse}, we do not restart the world-camera point correspondence search from scratch every time but instead keep a set number inliers from the past iteration (e.g., \SI{50}{\percent}). This helps stabilize the pose search and avoid large variations across iterations.
Finally, we limit the number of times \textsc{Stall} and \textsc{Reseed} can be performed consecutively and return a failure code if reached.

\section{Dataset}
\label{sec:dataset}

\begin{figure}[h]
    \centering
    \subfloat[Flagstone testbed.]{
        \includegraphics[height=0.32\columnwidth]{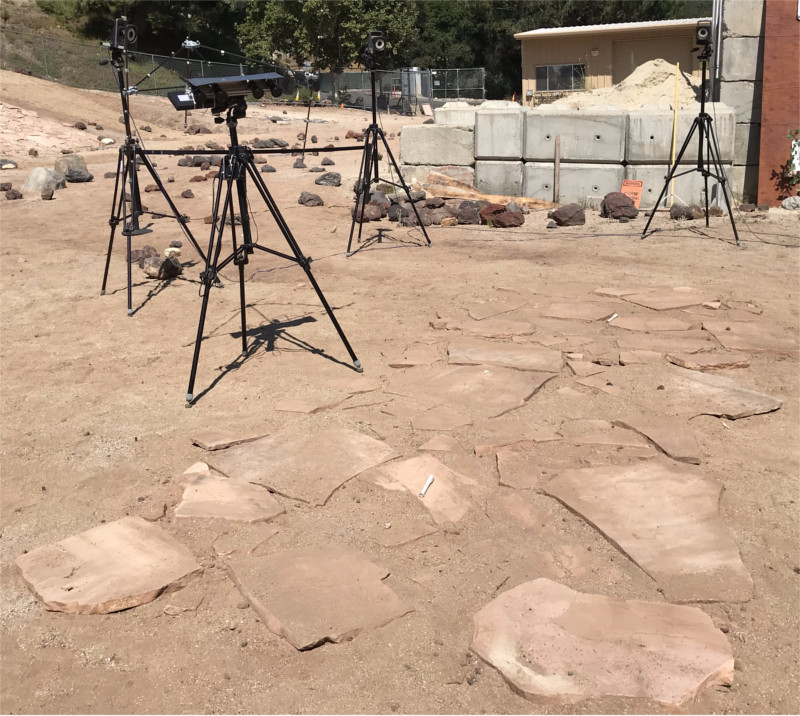}
        \label{fig:outdoordataset:testbed}
    }
    \subfloat[Camera pose grid around tubes $\Mtube_A$, $\Mtube_B$.]{
        \resizebox{!}{0.32\columnwidth}{
            \Mtikzfigoutdoordataset
        }
        \label{fig:outdoordataset:grid}
    }
    \\
    \subfloat[Sample images on CFA 6 rocks (left) and CFA 2 pebbles (right).]{
        \includegraphics[width=0.99\columnwidth]{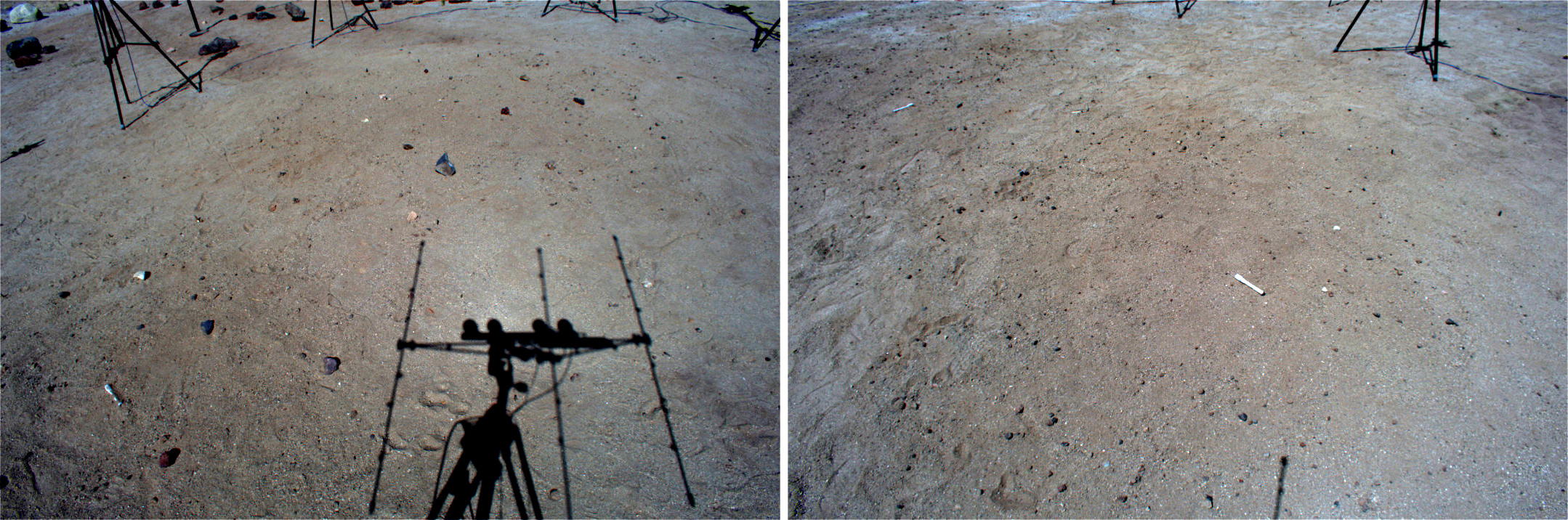}
        \label{fig:outdoordataset:sample}
    }
    \caption{Outdoor dataset: (a) testbed, (b) capture grid, (c) sample images.}
    \label{fig:outdoordataset}
\end{figure}

To evaluate the performance of our proposed method towards mission planning, we collect an outdoor dataset capturing representative conditions we expect to face on Mars.
In a preliminary study~\cite{aeroconf:pham:2020}, we showed on an indoor dataset that lighting changes and lack of consistent features throughout were a major challenge for rover localization.
We investigate these problems further by collecting a new dataset, outdoor, enabling:
\begin{enumerate*}
    \item natural light and shadows from the sun that change continuously;
    \item diverse, representative terrains, based on discussions with Mars geologists on where sample tubes could be dropped;
    \item a large workspace to evaluate different depot imaging strategies as well as in-depot navigation during tube recovery (e.g., how far SFR can deviate from areas imaged by Perseverance).
\end{enumerate*}

We construct a camera acquisition setup for outdoor use consisting of four FLIR BlackFly S cameras  ($5472 \times 3648$, color, \SI{77}{\degree} field of view) arranged as two stereo pairs of baseline \SI{20}{} and \SI{40}{\centi\meter}, representative of the Perseverance rover's optics~\cite{lpsc:maki:2020}. The cameras are covered by an aluminum plate serving as heat shield for extended use under sunlight and rigidly linked to a frame carrying motion capture (MoCap) markers for pose ground-truthing using 10 Vicon T-160 cameras (see Fig.~\ref{fig:outdoordataset:testbed}).
The dataset captures the following:
\begin{itemize}
    \item 3 terrain types (see Figs.~\ref{fig:outdoordataset:testbed} and~\ref{fig:outdoordataset:sample}):
    \begin{itemize}
        \item \enquote{Flagstone}: broken stone slabs covered with a thin layer of sand similarly to fractured bedrock on Mars.
        \item \enquote{CFA6}: a rock distribution of \emph{cumulative fractional area} (CFA -- a measure of rock density~\cite{jgrp:golombek:1997}) equal to \SI{6}{\percent}, the smallest rocks still visible from orbit to guide the choice of depot location. Rocks encountered in practice would only be this big or smaller.
        \item \enquote{CFA2}: small rocks only visible from surface imagery (not orbit), here pebbles on dust and sand.
    \end{itemize}
    \item 3 image capture times: \enquote{am} (9 am to 10 am), \enquote{nn} (noon to 1 pm), \enquote{pm} (3 pm to 4 pm)
    \item 2 sample tubes with variable visibility: unoccluded or in a crack between slabs (flagstone), unoccluded or \SI{25}{\percent} covered by sand (CFA 2), \SI{50}{} or \SI{75}{\percent} occluded (CFA6)
    \item 15 camera tripod positions along two circles, each centered on a sample tube (30 stops total, see Fig.~\ref{fig:outdoordataset:grid})
    \item 2 camera orientations at each stop (look at each tube)
    \item 4 cameras: 2 stereo pairs of baseline \SI{20}{} and \SI{40}{\centi\meter}
\end{itemize}

Overall, our outdoor dataset comprises $540$ capture configurations, totalling $2160$ images collected over the course of 3 days, annotated with reference poses from MoCap.

\section{Experiments}
\label{sec:experiments}

In this Section, we examine the performance of our method on different types of terrains and lighting conditions, then assess different depot mapping and navigation strategies to make recommendations for Mars Sample Return planning.

\subsection{Relocalization Performance}
\label{sec:experiments:nominal}

The first step consists in building a map of the depot.
For each terrain and capture time (9 combinations total), we build a depot map from 60 viewpoints captured with the \SI{40}{\centi\meter}-baseline stereo cameras, representative of those Perseverance would use for depot imaging.
Using the Agisoft Metashape software, we generate a textured mesh $\Mmesh$ of the full scene.
For Flagstone and CFA6 (possible scene perturbations, e.g., from Martian wind), we manually process $\Mmesh$ with the Blender 3D graphics software to only keep rocks in the sampling mask $\Mmeshfilter$.
For CFA2 (undisturbed scene), we keep the full mesh and set $\Mmeshfilter := \Mmesh$.
We depict the resulting meshes and masks in Figs.~\ref{fig:experiments:outdoor:mesh:fs} to~\ref{fig:experiments:outdoor:mesh:c2}.
We then use this map to relocalize images taken at different times of the day, see Fig.~\ref{fig:experiments:outdoor:map}.
Viewpoints to relocalize (green arrows) overlap with mapping images (blue) when they are from the same capture time, otherwise vary slightly over different capture sessions.

We evaluate our method by starting away from the ground-truth pose $\Mtransform{\Mworld}{\Mfetch}$, applying a random transformation of \SI{10}{}-to-\SI{20}{\centi\meter} translation and \SI{1.5}{\degree} rotation components (see Section~\ref{sec:method:overview}).
We run VTSM using the resulting perturbed pose $\Mtransformhat{\Mworld}{\Mfetch}$ as initial guess, $\Mtranslationuncertainty_0 = \SI{20}{\centi\meter}$ and $\Mrotationuncertainty_0 = \SI{1.5}{\degree}$ initial search bounds with decay parameter $\Mdecay = 0.5$, virtual template sizes $128$ and $256$, and epipolar constraint threshold $\Mmatchthreshold = \SI{8}{pixels}$.
A run is considered successful if $\Mcorrespondencescount = 100$ correspondences are found and the pose is successfully updated for $\Miterationcount = 5$ iterations.
The transformation estimation error is defined as the difference between ground-truth pose and final estimate, $\Mtransformhat{\Mfetch}{\Mfetch} = (\Mtransform{\Mworld}{\Mfetch})^{-1} \cdot \Mtransformhat{\Mworld}{\Mfetch}$.

We report in Fig.~\ref{fig:experiments:outdoor:results:all} the VTSM relocalization success rate and average error on each terrain type, sorted by time difference between mapping and relocalization: \SI{0}{\hour} (same time for both), \SI{3}{\hour} (e.g., noon relocalization vs morning map), \SI{6}{\hour} (e.g., afternoon relocalization vs morning map).
We also depict Flagstone detailed results in Fig.~\ref{fig:experiments:outdoor:results:fs} as 2D plots where each point's coordinates represent the distance (linear and angular) between the viewpoint to relocalize and the nearest viewpoint used to map the depot, and its color the relocalization accuracy.
As a baseline to our method, we also report these metrics when performing localization by matching local features between the same relocalization and nearest mapping viewpoints.
We did so using the LIBVISO2 package~\cite{iv:geiger:2011}, modified to use SIFT features for better robustness in exchange for longer computation time.

We observe the following.
First, while SIFT-based localization is generally more accurate than VTSM when it successfully estimates a transformation, this success rate decreases significantly with lighting changes across all terrains.
Notably, it completely drops to \SI{0}{\percent} with \SI{6}{\hour} of natural sunlight difference while VTSM achieves \SI{100}{\percent} across all configurations.
We note that relocalization errors are largest on CFA6, possibly due to the small size of rock features together with their sparsity in the sampling mask (see Fig.~\ref{fig:experiments:outdoor:mesh:c6}), suggesting that it may be preferrable to build the depot on large-enough rocks if the scene is expected to be affected by Martian wind (Flagstone), or to find areas where sand will remain undisturbed (CFA2).
Second, Fig.~\ref{fig:experiments:outdoor:results:fs} illustrates that VTSM is generally successful throughout the assessed range of \SI{40}{\centi\meter} and \SI{20}{\degree} between test and mapping viewpoints. This prompts us to consider larger ranges in the next sections.
Finally, we report an average run time of \SI{3}{\minute} \SI{55}{\second} on a \SI{4.5}{\giga\hertz} CPU-only, single-threaded implementation compatible with space-rated hardware.

For the sake of completeness, we considered alternative methods that may \emph{not} be compatible with space-rated hardware, such as neural-network-based features~\cite{iccv:simo:2015}, that we observed generally performing worse than SIFT on Mars-like terrains under lighting changes~\cite{aeroconf:pham:2020}.
We also implemented the image transform of~\cite{iros:corke:2013} as a preprocessing step and observed that the added noise, as reported by~\cite{icra:paton:2015}, particularly hindered subsequent local feature matching on our relatively barren terrains.
While this may be alleviated with additional training, it remains unclear whether the single Planckian source assumption would hold at tube pickup time on Mars.
Finally, we trained the benchmark-leading~\cite{cvpr:sattler:2018} image retrieval technique of~\cite{cvpr:torii:2015} on our dataset.
While image retrieval itself aims at returning \emph{the nearest pose in a predefined database} rather than the actual rover pose itself, we could envision using such a system to seed VTSM with a coarse estimate for further refinement.
However, we again observed the method to fail at retrieving such a pose, which may be due to the DenseVLAD features employed being derived from SIFT, therefore subject to similar limitations.

\newcommand{\Mtableoutdoorfs}{
    \setlength{\tabcolsep}{0.15em} % for the horizontal padding
    \begin{tabular}{|c|c|c|c|}
        \hline
        & \SI{0}{\hour} difference & \SI{3}{\hour} difference & \SI{6}{\hour} difference \\
        \hline
        \parbox[t]{2mm}{\multirow{7}{*}{\rotatebox[origin=c]{90}{SIFT}}} &
        \multirow{6}{*}{\includegraphics[width=0.30\columnwidth]{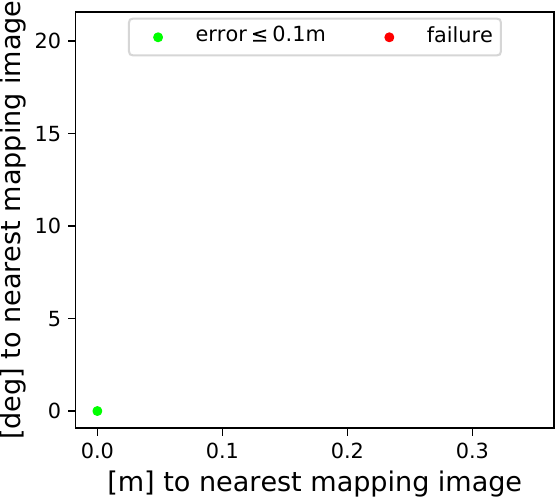}} &
        \multirow{6}{*}{\includegraphics[width=0.30\columnwidth]{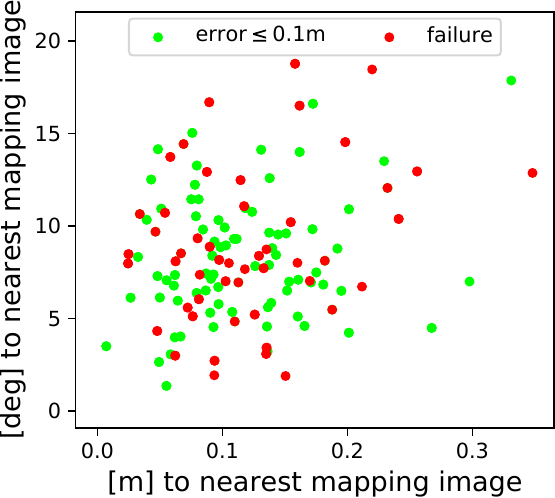}} &
        \multirow{6}{*}{\includegraphics[width=0.30\columnwidth]{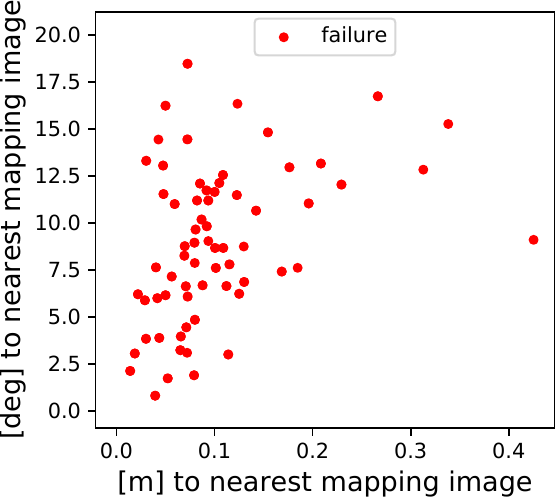}} \\
        & & & \\
        & & & \\
        & & & \\
        & & & \\
        & & & \\
        \cline{2-4}
        & \small{\SI{98.9}{\percent}, \SI{0.8}{\milli\meter}} & \small{\SI{66.7}{\percent}, \SI{22.8}{\milli\meter}} & \small{\SI{0.0}{\percent}, N/A} \\
        \hline
        \parbox[t]{2mm}{\multirow{7}{*}{\rotatebox[origin=c]{90}{VTSM}}} & 
        \multirow{6}{*}{\includegraphics[width=0.30\columnwidth]{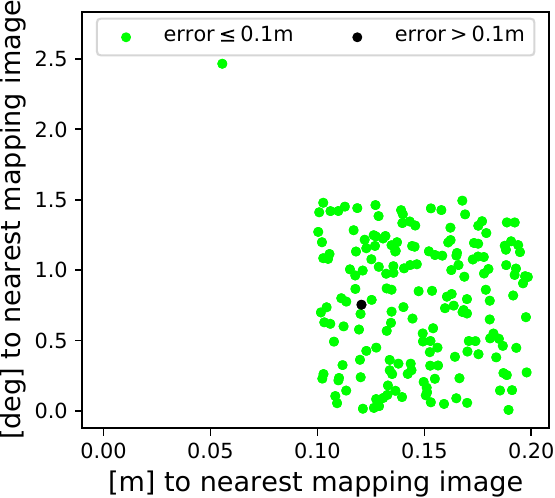}} &
        \multirow{6}{*}{\includegraphics[width=0.30\columnwidth]{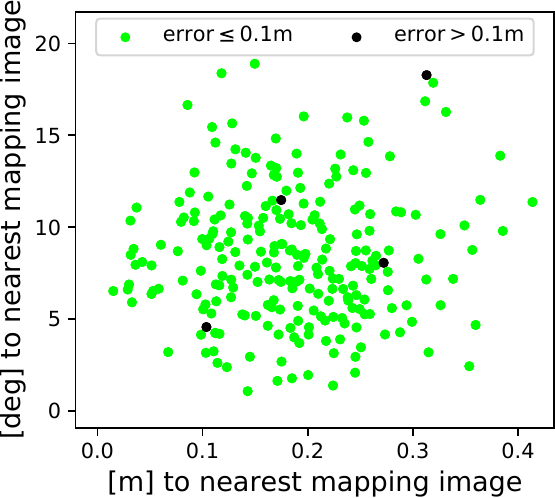}} &
        \multirow{6}{*}{\includegraphics[width=0.30\columnwidth]{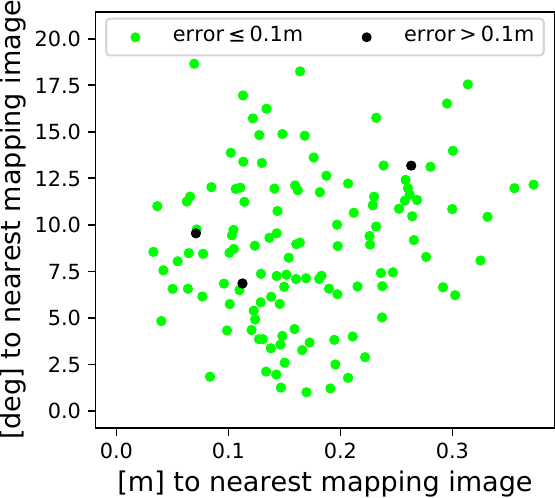}} \\
        & & & \\
        & & & \\
        & & & \\
        & & & \\
        & & & \\
        \cline{2-4}
        & \small{\SI{100}{\percent}, \SI{51.2}{\milli\meter}} & \small{\SI{100}{\percent}, \SI{56.0}{\milli\meter}} & \small{\SI{100.0}{\percent}, \SI{50.2}{\milli\meter}} \\
        \hline
    \end{tabular}
}

\newcommand{\Mtableoutdoorall}{
    \setlength{\tabcolsep}{0.30em} % for the horizontal padding
    \begin{tabular}{|c|r|ccc|ccc|ccc|}
        \hline
        \multicolumn{2}{|r|}{Terrain} & \multicolumn{3}{c|}{Flagstone} & \multicolumn{3}{c|}{CFA6} & \multicolumn{3}{c|}{CFA2} \\
        \hline
        \multicolumn{2}{|r|}{Time diff. [\SI{}{\hour}]} & \SI{0}{} & \SI{3}{} & \SI{6}{} & \SI{0}{} & \SI{3}{} & \SI{6}{} & \SI{0}{} & \SI{3}{} & \SI{6}{}  \\
        \hline
        \parbox[t]{2mm}{\multirow{2}{*}{\rotatebox[origin=c]{90}{SIFT}}} & Success [\SI{}{\percent}] &
        98.9&66.7&0.0 &
        100.0&70.7&0.0 &
        98.9&52.9&0.0 \\
        & Error [\SI{}{\milli\meter}] &
        0.8&22.8& N/A &
        0.4&6.5& N/A &
        0.6&10.7& N/A \\
        \hline
        \parbox[t]{2mm}{\multirow{3}{*}{\rotatebox[origin=c]{90}{VTSM}}} & Success [\SI{}{\percent}] &
        100.0&100.0&100.0 &
        100.0&100.0&100.0 &
        100.0&100.0&100.0 \\
        & Error (init) [\SI{}{\milli\meter}] &
        148.1&152.3&149.2 &
        149.7&147.9&150.0 &
        149.3&150.5&148.0 \\
        & Error (end) [\SI{}{\milli\meter}] &
        51.2&56.0&50.2 &
        65.4&75.9&98.5 &
        51.1&52.8&76.6 \\
        \hline
    \end{tabular}
}

\begin{figure}[h]
    \centering
    \subfloat[Mapping and test viewpoints.]{
        \includegraphics[height=0.27\columnwidth]{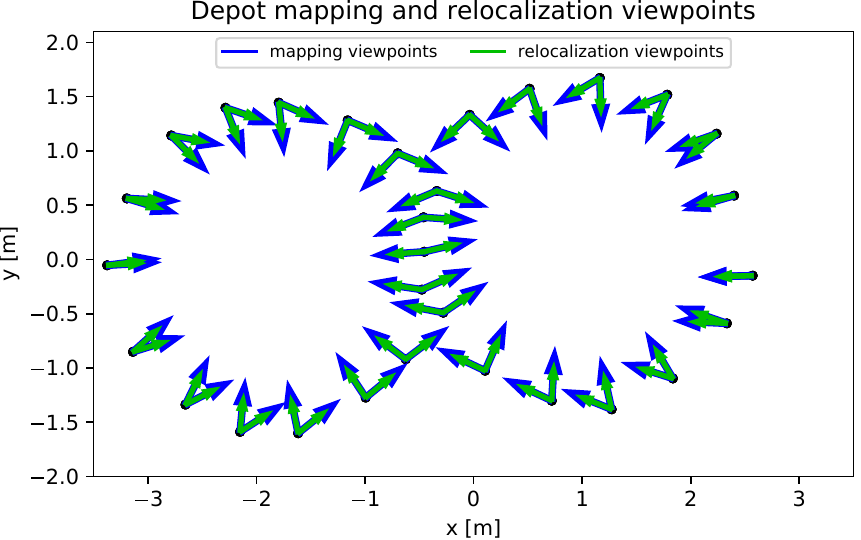}
        \label{fig:experiments:outdoor:map}
    }
    \subfloat[Flagstone mesh $\Mmesh$ (green: mask $\Mmeshfilter$).]{
        \includegraphics[height=0.27\columnwidth]{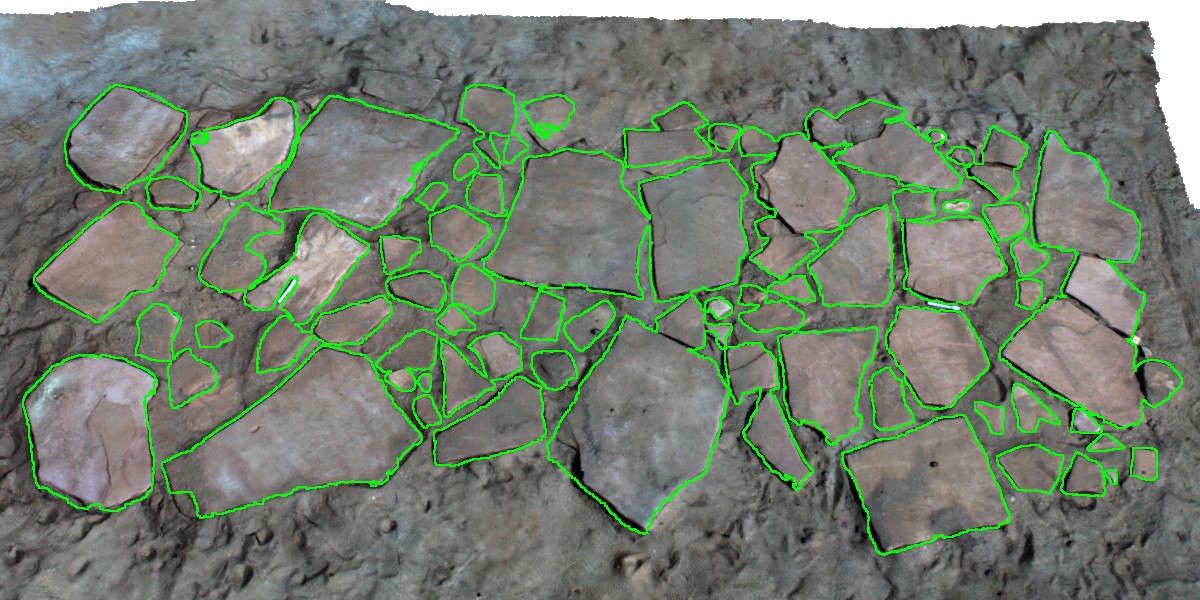}
        \label{fig:experiments:outdoor:mesh:fs}
    }
    \\
    \subfloat[CFA6 mesh $\Mmesh$ (green: mask $\Mmeshfilter$).]{
        \includegraphics[height=0.245\columnwidth]{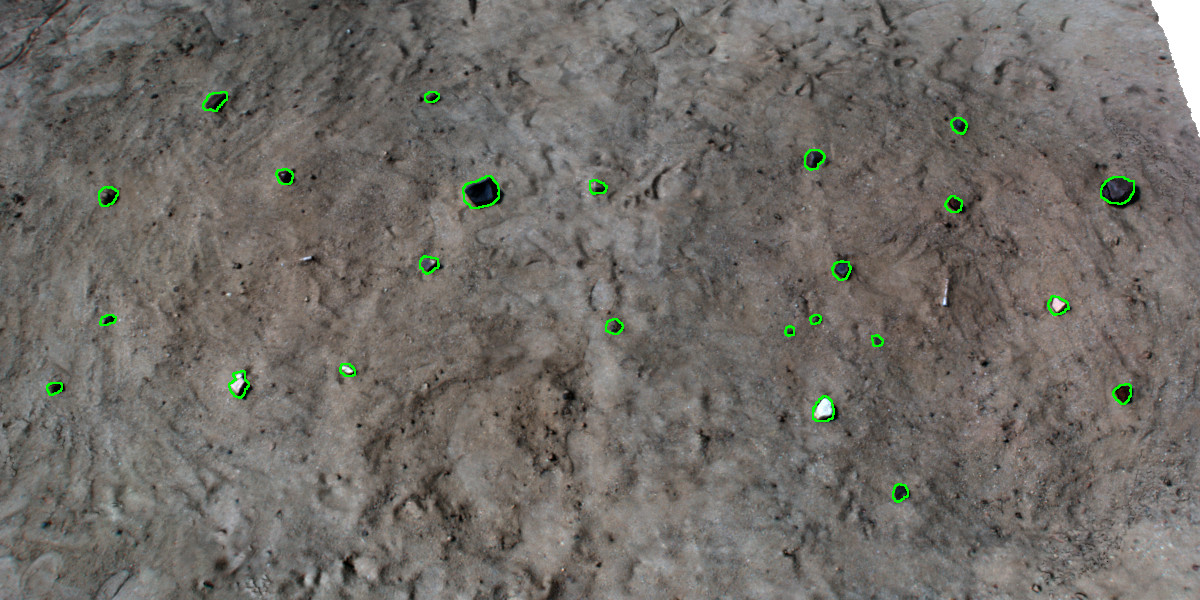}
        \label{fig:experiments:outdoor:mesh:c6}
    }
    \subfloat[CFA2 mesh $\Mmesh$ (no mask $\Mmeshfilter$).]{
        \includegraphics[height=0.245\columnwidth]{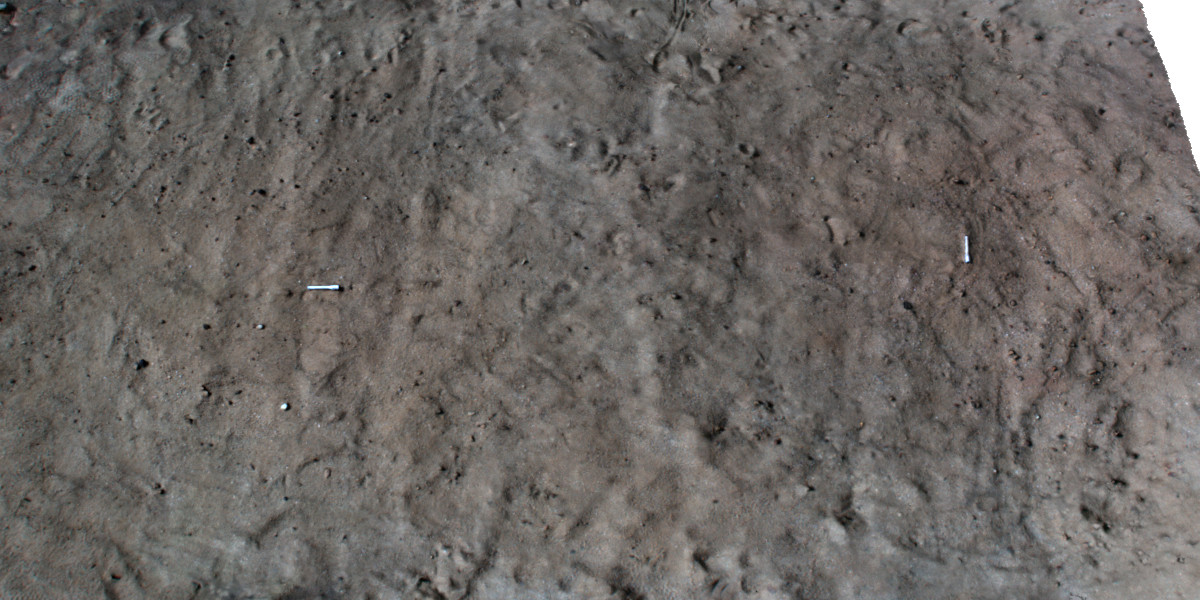}
        \label{fig:experiments:outdoor:mesh:c2}
    }
    \\
    \subfloat[Localization success rate and average error on all terrain-time differences.]{
        \resizebox{1.0\columnwidth}{!}{
            \Mtableoutdoorall
        }
        \label{fig:experiments:outdoor:results:all}
    }
    \\
    \subfloat[Transformation estimation success rate and average error on flagstone depot across \SI{0}{}, \SI{3}{}, \SI{6}{\hour} capture time and lighting differences. Note: SIFT-\SI{0}{\hour} amounts to estimating the (zero) transformation between the same images and VTSM-\SI{0}{\hour} illustrates the effects of \SI{10}{}-to-\SI{20}{\centi\meter}, \SI{0}{}-to-\SI{1.5}{\degree} pose randomization alone. \SI{3}{} and \SI{6}{\hour} plots include pose variations across different depot traverses.]{
        \Mtableoutdoorfs
        \label{fig:experiments:outdoor:results:fs}
    }
    \caption{VTSM evaluation on all terrain-time differences.}
    \label{fig:experiments:outdoor}
\end{figure}

\subsection{Parameter Sensitivity Analysis}
\label{sec:experiments:alt}

\newcommand{\Mtableoutdooralt}{
    \setlength{\tabcolsep}{0.30em} % for the horizontal padding
    \begin{tabular}{|c|r|ccc|ccc|ccc|}
        \hline
        \multicolumn{2}{|r|}{Terrain-time} & \multicolumn{3}{c|}{Flagstone-\SI{6}{\hour}} & \multicolumn{3}{c|}{CFA6-\SI{6}{\hour}} & \multicolumn{3}{c|}{CFA2-\SI{6}{\hour}} \\
        \hline
        \multicolumn{2}{|r|}{Configuration} & \textbf{ref} & \textbf{bl} & \textbf{50cm} & \textbf{ref} & \textbf{bl} & \textbf{50cm} & \textbf{ref} & \textbf{bl} & \textbf{50cm} \\
        \hline
        \parbox[t]{2mm}{\multirow{3}{*}{\rotatebox[origin=c]{90}{VTSM}}} & Success [\SI{}{\percent}] & 
        100.0 & 100.0 & 100.0 &
        100.0 & 100.0 & 100.0 &
        100.0 & 100.0 & 100.0  \\
        & Error (init) [\SI{}{\milli\meter}] &
        149.2 & 146.2 & 363.8 &
        150.0 & 153.7 & 379.2 &
        148.0 & 146.5 & 377.1 \\
        & Error (end) [\SI{}{\milli\meter}] &
        50.2 & 58.2  & 49.3 &
        98.5 & 123.0 & 83.5 &
        76.6 & 84.6  & 62.2 \\
        \hline
    \end{tabular}
}

\newcommand{\Mtablematchingresults}{
    \setlength{\tabcolsep}{0.15em} % for the horizontal padding
    \begin{tabular}{|c|c|c|c|}
        \hline
        & Synthetic, random & Real, relocalize & Synthetic, final \\
        \hline
        \parbox[t]{2mm}{\multirow{4}{*}{\rotatebox[origin=c]{90}{Flagstone}}} & 
        \multirow{4}{*}{\includegraphics[trim=0 0 0 10, clip, width=0.30\columnwidth]{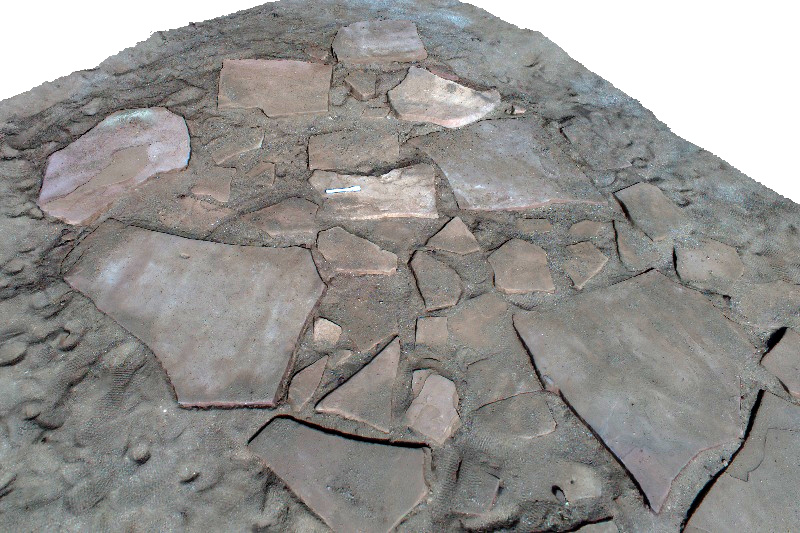}} &
        \multirow{4}{*}{\includegraphics[trim=0 0 0 10, clip, width=0.30\columnwidth]{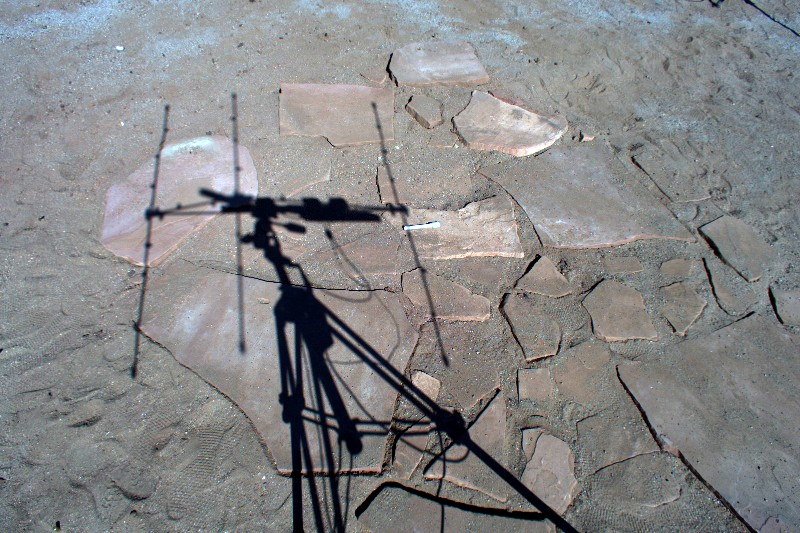}} &
        \multirow{4}{*}{\includegraphics[trim=0 0 0 10, clip, width=0.30\columnwidth]{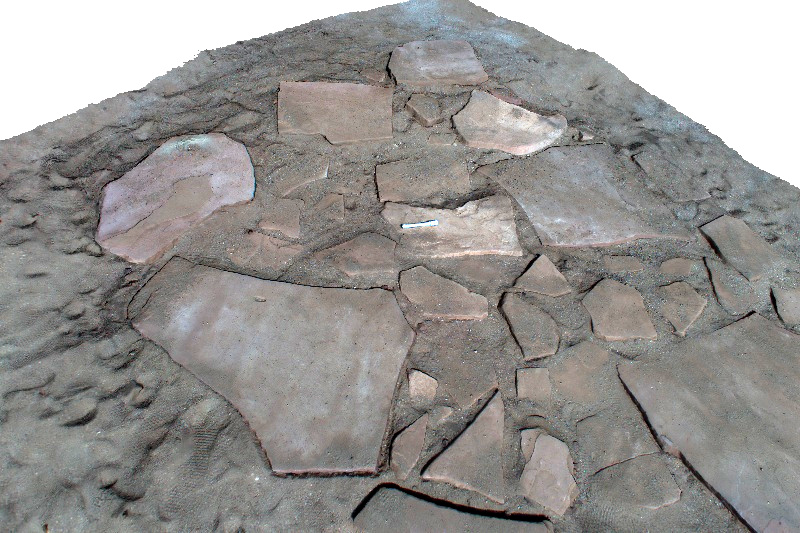}} \\
        & & & \\
        & & & \\
        & & & \\
        %& & & \\
        \hline
        \parbox[t]{2mm}{\multirow{4}{*}{\rotatebox[origin=c]{90}{CFA6}}} & 
        \multirow{4}{*}{\includegraphics[trim=0 0 0 10, clip, width=0.30\columnwidth]{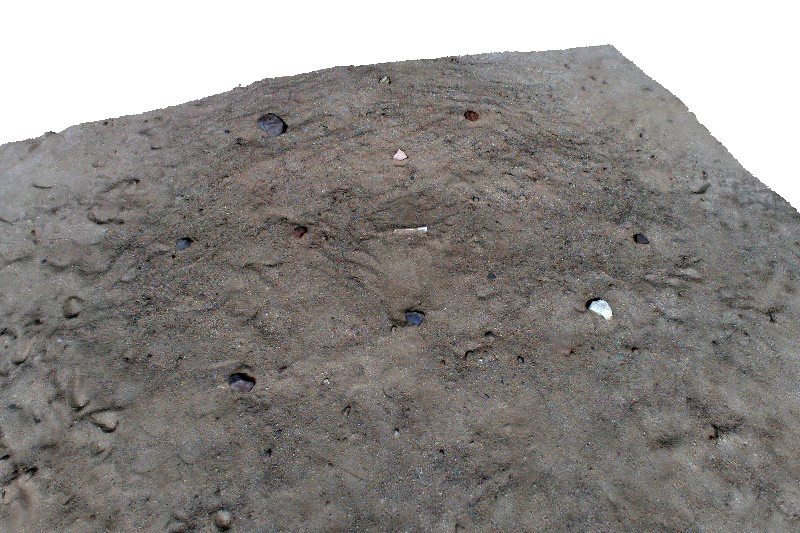}} &
        \multirow{4}{*}{\includegraphics[trim=0 0 0 10, clip, width=0.30\columnwidth]{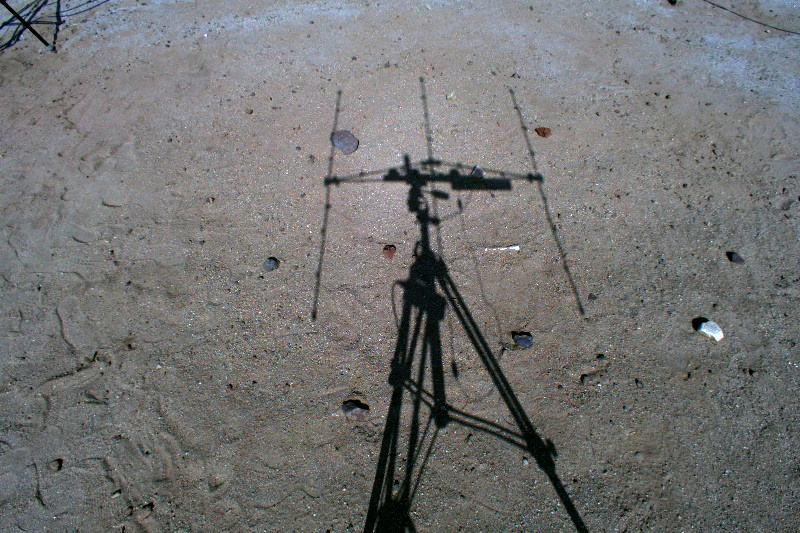}} &
        \multirow{4}{*}{\includegraphics[trim=0 0 0 10, clip, width=0.30\columnwidth]{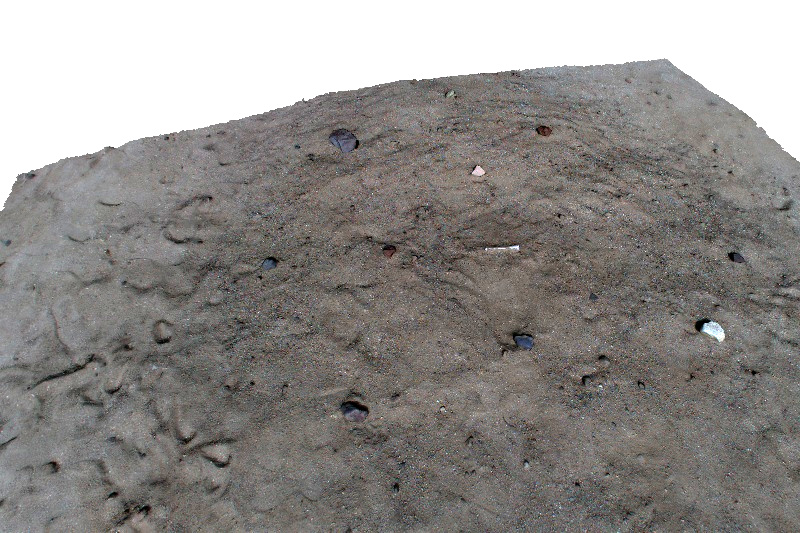}} \\
        & & & \\
        & & & \\
        & & & \\
        \hline
        \parbox[t]{2mm}{\multirow{4}{*}{\rotatebox[origin=c]{90}{CFA2}}} & 
        \multirow{4}{*}{\includegraphics[trim=0 0 0 10, clip, width=0.30\columnwidth]{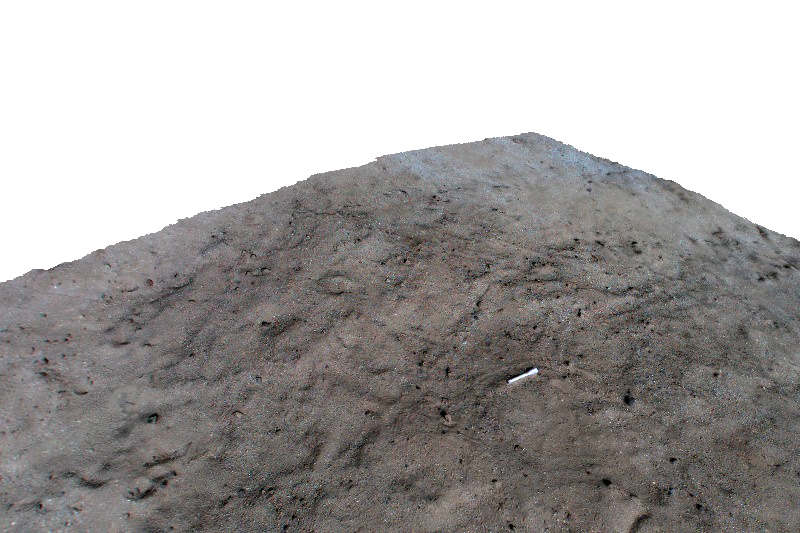}} &
        \multirow{4}{*}{\includegraphics[trim=0 0 0 10, clip, width=0.30\columnwidth]{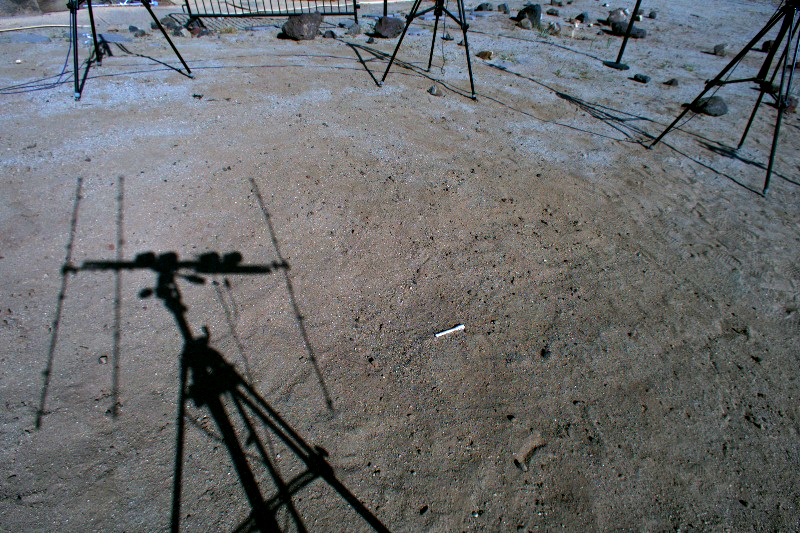}} &
        \multirow{4}{*}{\includegraphics[trim=0 0 0 10, clip, width=0.30\columnwidth]{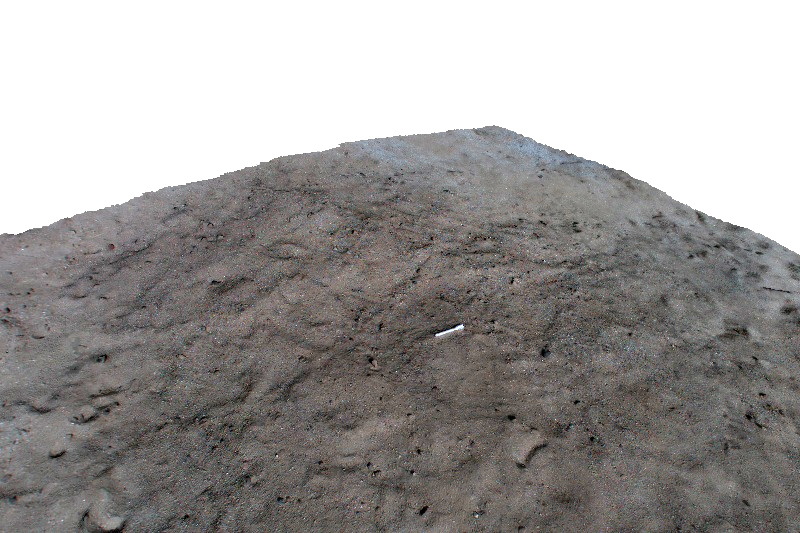}} \\
        & & & \\
        & & & \\
        & & & \\
        \hline
    \end{tabular}
}

\begin{figure}[h]
    \centering
    \subfloat[VTSM results on different configurations. Reference results \textbf{ref}: \SI{40}{\centi\meter} relocalization baseline, \SI{10}{}-to-\SI{20}{\centi\meter} initial randomization. Two variations: \textbf{bl}: \SI{20}{\centi\meter} relocalization baseline, \textbf{50cm}: \SI{25}{}-to-\SI{50}{\centi\meter} initial randomization.]{
        \resizebox{1.0\columnwidth}{!}{
            \Mtableoutdooralt
        }
        \label{fig:experiments:alternative:table}
    }
    \\
    \subfloat[Afternoon relocalization vs morning map with \SI{50}{\centi\meter} randomization. Note the real image strong shadows, which could occur on Mars (rover shadow).]{
        \resizebox{1.0\columnwidth}{!}{
            \Mtablematchingresults
        }
        \label{fig:experiments:alternative:matching}
    }
    \caption{Results on alternative relocalization configurations.}
    \label{fig:experiments:alternative}
\end{figure}

While the \SI{40}{\centi\meter}-baseline stereo cameras are representative of those Perseverance will use for depot mapping, the SFR design may be constrained by further size and payload requirements.
Keeping the depot maps built from the \SI{40}{\centi\meter}-baseline stereo images, we now relocalize images taken by other cameras in a \SI{20}{\centi\meter}-baseline configuration.
We report the resulting errors on all three terrains over \SI{6}{\hour} time difference between mapping and relocalization in Fig.~\ref{fig:experiments:alternative:table}.
We observe that while the smaller stereo baseline results in higher relocalization errors, these are expected as the theoretical depth uncertainty itself also increases by \SI{6.5}{}-to-\SI{26.0}{\milli\meter} for a \SI{1}{pixel} disparity uncertainty \SI{3}{}-to-\SI{6}{\meter} ahead of the camera.
As success rates remain similar, we infer here that VTSM itself is robust to camera changes, with its accuracy contingent on that permitted by the chosen setup.

We now consider an alternative scenario where the rover's position uncertainty suddenly grows beyond the previous \SI{20}{\centi\meter} to \SI{50}{\centi\meter} (e.g., slippage).
To address this, one possibility is to simply increase the randomization range $\Mtranslationuncertainty_0$ when generating synthetic viewpoints to \SI{50}{\centi\meter} around the current pose guess $\Mtransformhat{\Mworld}{\Mfetch}$.
However, the increased search range also requires more attempts to sample poses closer to the real one, while also generating more false positives.
Furthermore, template matching takes longer as the increased uncertainty only lets us restrict the search to about $2000 \times 2000$ patches within the real images instead of $800 \times 800$ as described in Section~\ref{sec:method:matching}.
Instead, we propose to keep VTSM virtual viewpoint randomization at \SI{20}{\centi\meter}, but around \emph{multiple pose seeds} randomly sampled within the \SI{50}{\centi\meter} uncertainty range.
We choose to run one VTSM iteration over $100$ such pose seeds and select the one resulting in the maximum number of correspondence inliers to run the rest of the algorithm on.
Adding these results to Fig.~\ref{fig:experiments:alternative:table}, we observe that success rates remain at \SI{100}{\percent} and that relocalization errors are slightly lower, which may stem from the search over multiple pose seeds within a \SI{50}{\centi\meter} range being a better initializer than relocalizing from a single one within \SI{20}{\centi\meter}.
Fig.~\ref{fig:experiments:alternative:matching} depicts real and synthetic images from VTSM showing successful matching on all terrains despite strong lighting changes between mapping and relocalization.
We report an average \SI{1}{\hour} \SI{4}{\minute} run time per relocalization attempt with \SI{50}{\centi\meter} uncertainty, including \SI{1}{\hour} to evaluate the $100$ pose seeds, which could be improved using other search schemes or early stop criteria (e.g., minimum inlier ratio).

\subsection{Depot Imaging Strategy and Relocalization Range}
\label{sec:experiments:subsets}

Finally, we consider the case where SFR deviates from the path Perseverance took when imaging the depot.
We do so in two ways.
First, we consider two alternative depot imaging trajectories for Perseverance (see Fig.~\ref{fig:experiments:subsets:trajectories}): a \emph{wave} trajectory where the rover navigates between two consecutive sample tubes imaging both, and a \emph{forward} trajectory where the rover only looks ahead while driving.
This lets us evaluate relocalization from viewpoints further from the mapping set.
Second, we sub-sample each depot imaging set by only using viewpoints every $x$ \SI{}{\meter}, which lets us evaluate Perseverance imaging density requirements to enable SFR relocalization.
We observe the following.
First, our meshing software starts failing to align input images when the imaging step size exceeds \SI{1}{\meter} due to insufficient overlap between viewpoints.
We depict in Fig.~\ref{fig:experiments:subsets:unaligned} the forward path with step size \SI{1.9}{\meter}, leaving only the first and last viewpoints of the trajectory as mapping images, which the software could not be align.
We thus obtained a mesh from only the first viewpoint's stereo pair for Flagstone and CFA6, and from the final viewpoint for CFA2. The latter being in the middle of the depot resulted in only half of it being 3D-modeled.
While this issue may be mitigated using further image alignment techniques, we are also interested in evaluating our method against less accurate meshes.
Fig.~\ref{fig:experiments:subsets:stepsizesix} illustrates that VTSM relocalization is still successful up to the \SI{6}{\meter} away from the nearest mapping image on Flagstone and CFA6, with errors mostly below \SI{10}{\centi\meter} up to \SI{3}{\meter} away.
Failure cases on CFA2 appear, expectedly, when attempting to relocalize images of the depot half that could not be mapped.
We report results on all configurations in Fig.~\ref{fig:experiments:subsets:all}, observing that relocalization accuracy drops after \SI{1}{\meter} depot imaging step in most cases.

\newcommand{\Mtablestepsizesix}{
    \setlength{\tabcolsep}{0.15em} % for the horizontal padding
    \begin{tabular}{|c|c|c|c|}
        \hline
        & Flagstone - \SI{6}{\hour} & CFA6 - \SI{6}{\hour} & CFA2 - \SI{6}{\hour} \\
        \hline
        \parbox[t]{2mm}{\multirow{7}{*}{\rotatebox[origin=c]{90}{Step size 6: \SI{1.9}{\meter}}}} & 
        \multirow{6}{*}{\includegraphics[width=0.30\columnwidth]{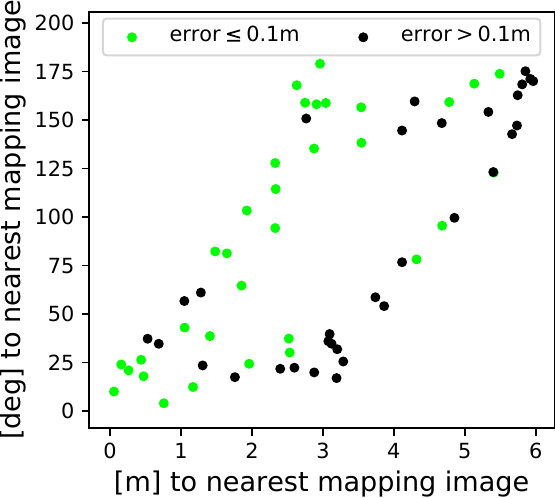}} &
        \multirow{6}{*}{\includegraphics[width=0.30\columnwidth]{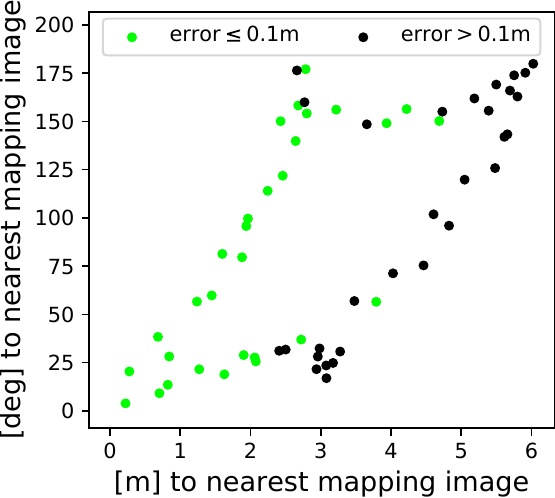}} &
        \multirow{6}{*}{\includegraphics[width=0.30\columnwidth]{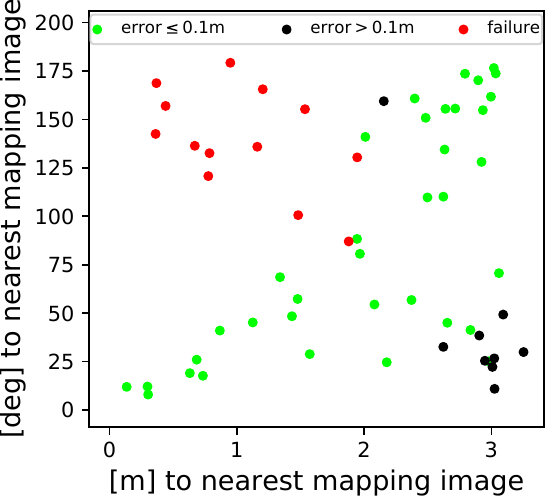}} \\
        & & & \\
        & & & \\
        & & & \\
        & & & \\
        & & & \\
        \cline{2-4}
        & \small{\SI{100.0}{\percent}, \SI{104.2}{\milli\meter}} & \small{\SI{100.0}{\percent}, \SI{121.6}{\milli\meter}} & \small{\SI{77.6}{\percent}, \SI{84.4}{\milli\meter}} \\
        \hline
    \end{tabular}
}

\newcommand{\Mtablesubsetswaveforward}{
    \setlength{\tabcolsep}{0.30em} % for the horizontal padding
    \begin{tabular}{|c|r|cccc|cccc|}
        \hline
        \multicolumn{2}{|r|}{Trajectory} & \multicolumn{4}{c|}{Wave} & \multicolumn{4}{c|}{Forward} \\
        \hline
        \multicolumn{2}{|r|}{Step size [m]} & 
        0.4 & 0.8 & 1.5 & 2.0 &
        0.4 & 0.8 & 1.2 & 1.9 \\
        \hline
        \parbox[t]{2mm}{\multirow{3}{*}{\rotatebox[origin=c]{90}{F.stone}}} & Range [\SI{}{\meter}] & 
        3.0 & 3.0 & 3.0 & 6.0 &
        3.0 & 3.0 & 5.0 & 6.0 \\
        & Success [\SI{}{\percent}] &
        100.0 & 100.0 & 100.0 & 86.2 &
        100.0 & 100.0 & 100.0 & 100.0 \\
        & Error [\SI{}{\milli\meter}] &
        56.9 & 50.8 & 65.9 & 248.2 &
        80.4 & 84.9 & 116.7 & 104.2 \\
        \hline
        \parbox[t]{2mm}{\multirow{3}{*}{\rotatebox[origin=c]{90}{CFA6}}} & Range [\SI{}{\meter}] & 
        3.0 & 3.0 & 3.5 & 6.0 &
        4.0 & 4.0 & 4.0 & 6.0 \\
        & Success [\SI{}{\percent}] &
        100.0 & 100.0 & 100.0 & 100.0 &
        100.0 & 100.0 & 98.3 & 100.0 \\
        & Error [\SI{}{\milli\meter}] &
        85.2 & 83.9 & 122.4 & 119.2 &
        75.6 & 110.9 & 139.3 & 121.6 \\
        \hline
        \parbox[t]{2mm}{\multirow{3}{*}{\rotatebox[origin=c]{90}{CFA2}}} & Range [\SI{}{\meter}] & 
        2.5 & 3.0 & 4.5 & 6.0 &
        3.0 & 3.0 & 5.0 & 3.5 \\
        & Success [\SI{}{\percent}] &
        100.0 & 100.0 & 100.0 & 77.6 &
        100.0 & 100.0 & 100.0 & 77.6 \\
        & Error [\SI{}{\milli\meter}] &
        60.9 & 77.3 & 189.1 & 312.8 &
        74.3 & 75.1 & 94.4 & 84.4 \\
        \hline
    \end{tabular}
}

\begin{figure}[h]
    \centering
    \subfloat[Alternative imaging trajectories: wave (left) and forward (right).]{
        \includegraphics[width=0.49\columnwidth]{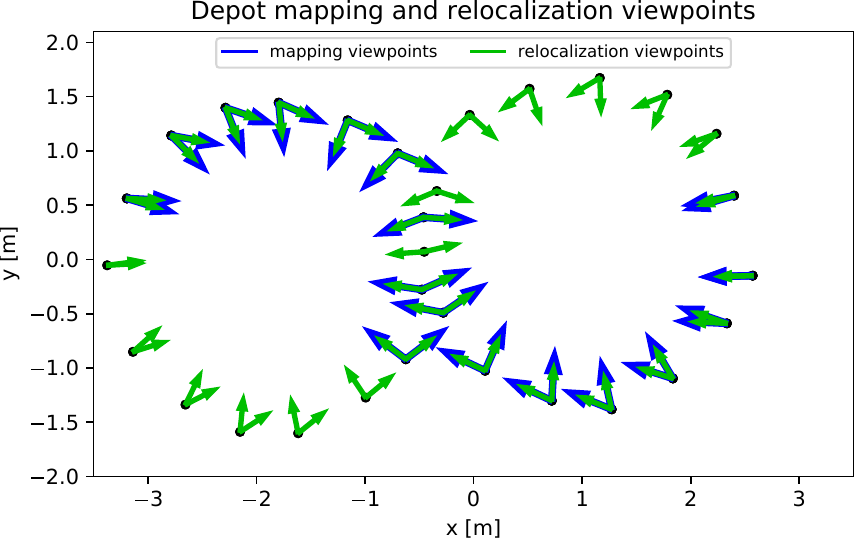}
        \includegraphics[width=0.49\columnwidth]{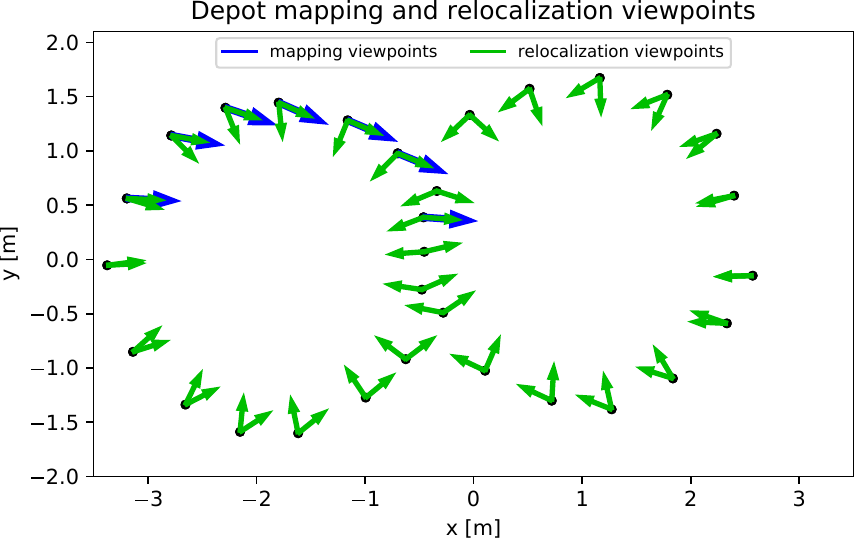}
        \label{fig:experiments:subsets:trajectories}
    }
    \\
    \subfloat[Forward trajectory with step size 6 (\SI{1.9}{\meter} step). Red arrows indicate mapping images that were not successfully aligned with the others (blue arrows) during mesh reconstruction. Left: Flagstone and CFA6, right: CFA2.]{
        \includegraphics[width=0.49\columnwidth]{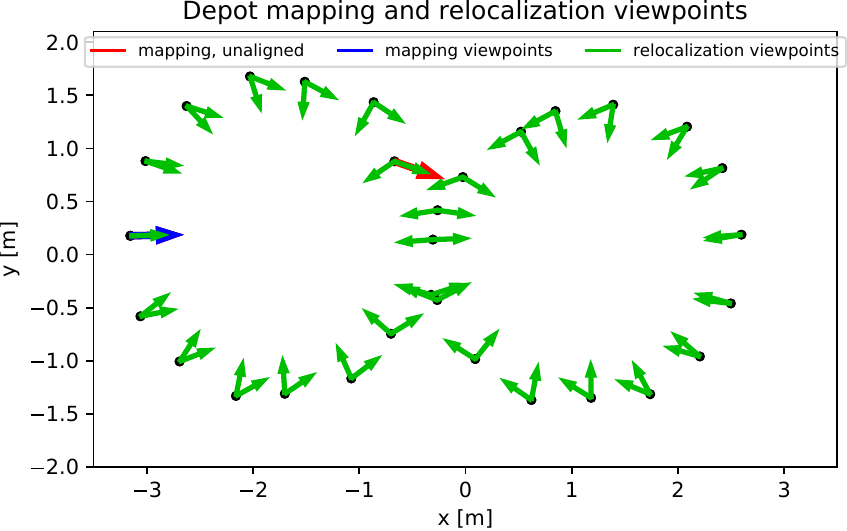}
        \includegraphics[width=0.49\columnwidth]{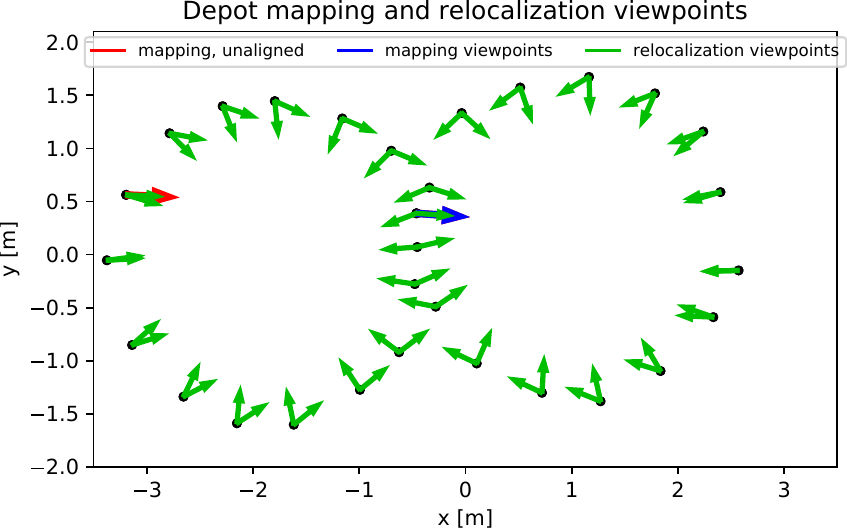}
        \label{fig:experiments:subsets:unaligned}
    }
    \\
    \subfloat[Transformation estimation success rate and average error on forward trajectory with step size 6. All maps are effectively built from a single stereo pair. Failure cases for CFA2 can be attributed to half the mesh missing.]{
        \resizebox{1.0\columnwidth}{!}{
            \Mtablestepsizesix
        }
        \label{fig:experiments:subsets:stepsizesix}
    }
    \\
    \subfloat[VTSM results on wave and forward imaging trajectories. We sub-sample each with four step sizes, skipping viewpoints in between. \enquote{Range} denotes the maximum distance between the resulting mapping and relocalization images.]{
        \resizebox{1.0\columnwidth}{!}{
            \Mtablesubsetswaveforward
        }
        \label{fig:experiments:subsets:all}
    }
    \caption{Results on different mapping step sizes and relocalization ranges.}
    \label{fig:experiments:subsets}
\end{figure}

\section{Discussion}
\label{sec:conclusion}

The problem of retrieving sample tubes on Mars years after they have been dropped by another rover is a difficult task due to the unknown of how depots may change over time.
In this paper, we presented a complete relocalization pipeline matching partial renderings of a depot map over multiple virtual viewpoints to real images.
Our approach estimated poses with \SI{100}{\percent} success rate across all terrains and lighting differences when local feature matching would completely fail, with average error below \SI{10}{\centi\meter} in both nominal and extended conditions.
Further analysis permitted by our large-scale dataset showed that our method maintained similar performance for at least \SI{3}{\meter} away from poses imaged by Perseverance, with relocalization failing only when depot mapping itself fails.
Based on results across different experimental conditions, we recommend that depots are constructed on fractured bedrock on Mars akin to flagstone on Earth, and imaged by Perseverance from viewpoints no further than \SI{1}{\meter} apart.
Sparse, or even no rocks can be considered if the effects of Martian wind can be deemed negligible from surface or orbital imagery.

Our work lends itself to multiple development opportunities.
First, the viewpoint randomization process could appropriately be implemented as a particle filter that updates search parameters rather than following a fixed schedule.
We expect this would improve relocalization accuracy and computational efficiency towards being used onboard the rover.
Machine learning techniques could also be used to identify salient points on the map that are most likely to yield successful matches, as done manually in~\cite{aeroconf:lorenz:2017}.
As a longer-term development, the synthetic matching pipeline could be applied to image modalities other than direct pixel intensity and its derivatives.
We could, for example, convert both synthetic and real images to lighting-invariant representations using recent neural-network-based techniques~\cite{ral:vonstumberg:2020}, or synthetically relight the depot map on the fly to reproduce Mars lighting conditions at SFR relocalization time.

\bibliographystyle{abbrv}
\bibliography{bib/paper}

\end{document}